\documentclass[10pt,twocolumn,letterpaper]{article}

\usepackage{cvpr}
\usepackage{times}
\usepackage{epsfig}
\usepackage{graphicx}
\usepackage{amsmath}
\usepackage{amssymb}
\usepackage{algorithm}
\usepackage{algpseudocode}
\usepackage{mathrsfs}
\usepackage{subcaption}
\usepackage{float}
\usepackage{dblfloatfix}
\usepackage[bottom]{footmisc}

\DeclareMathOperator*{\argmax}{arg\,max}

\usepackage[
  colorlinks=true,
]{hyperref}

\AtBeginDocument{
  \hypersetup{
    urlcolor=red,
    citecolor=green,
    linkcolor=black,
  }%
}

\def\chapterautorefname~#1\null{Chap.~#1\null}
\def\sectionautorefname~#1\null{Sec.~#1\null}
\def\subsectionautorefname~#1\null{Sec.~#1\null}
\def\figureautorefname~#1\null{Fig.~#1\null}
\def\tableautorefname~#1\null{Table.~#1\null}
\def\equationautorefname~#1\null{Eqn.~#1\null}
\def\algorithmautorefname~#1\null{Algo.~#1\null}

\renewcommand{\thefootnote}{\fnsymbol{footnote}}

\newcommand{\customfootnotetext}[2]{{
  \renewcommand{\thefootnote}{#1}
  \footnotetext[0]{#2}}}

\cvprfinalcopy 

\def\cvprPaperID{****} 

\begin{document}

\title{Universal Adversarial Perturbations: A Survey}

\author{Ashutosh Chaubey\textsuperscript{1} \qquad Nikhil Agrawal\textsuperscript{1} \qquad Kavya Barnwal\textsuperscript{2} \qquad Keerat K. Guliani\textsuperscript{2} \qquad Pramod Mehta\textsuperscript{2}\\ \\
Vision and Language Group\\ Indian Institute of Technology,
Roorkee\\
{\tt\small vlgiitr.github.io}}

\maketitle

\begin{abstract}
Over the past decade, Deep Learning has emerged as a useful and efficient tool to solve a wide variety of complex learning problems ranging from image classification to human pose estimation, which is challenging to solve using statistical machine learning algorithms. However, despite their superior performance, deep neural networks are susceptible to adversarial perturbations, which can cause the network's prediction to change without making perceptible changes to the input image, thus creating severe security issues at the time of deployment of such systems. Recent works have shown the existence of Universal Adversarial Perturbations, which, when added to any image in a dataset, misclassifies it when passed through a target model. Such perturbations are more practical to deploy since there is minimal computation done during the actual attack. Several techniques have also been proposed to defend the neural networks against these perturbations. In this paper, we attempt to provide a detailed discussion on the various data-driven and data-independent methods for generating universal perturbations, along with measures to defend against such perturbations. We also cover the applications of such universal perturbations in various deep learning tasks.
\end{abstract}

\section{Introduction}

\customfootnotetext{1}{\emph{denotes first authors with an equal contribution, in alphabetical order.}}
\customfootnotetext{2}{\emph{denotes second authors with an equal contribution, in alphabetical order.}}

Since the introduction of deep neural networks in effectively solving the ILSVRC~\cite{ILSVRC} image classification task~\cite{alexnet}, deep learning has expanded its horizons in solving a large number of complex tasks. The ability of deep neural networks to act as good function approximators has enabled them to successfully solve tasks such as image classification~\cite{googlenet,resnet,densenet}, object detection~\cite{faster_rcnn,yolo_v3}, instance segmentation~\cite{mask_rcnn}, language modeling~\cite{transformer,bert}, speech generation~\cite{wavenet,deepvoice3}, human pose estimation~\cite{lcrnet++,videopose3d}, etc. Due to this widespread use of deep learning techniques, the efficiency and reliability of these techniques become equally important when they are deployed in the real world. Many of these applications are crucial, requiring special focus on the safety and security of such systems~\cite{evtimov2017robust}.

Recent studies have shown that deep learning models are susceptible to some carefully constructed small noise called \textit{adversarial perturbations}, which when added to an input image, cause the network output to change drastically without making perceptible changes to the input image~\cite{intr_prop_of_nn}. Such perturbations pose a severe threat to the real-world application of any neural network model, especially in scenarios of autonomous vehicles and facial verification systems. Studies have shown that if an attacker has access to the model which needs to be attacked (i.e., the target model) along with access to only a few images from the distribution of images using which the target model has been trained, then they might achieve very high fooling rates (see \autoref{subsec:terminology}) on the entire distribution of images in the target dataset~\cite{intr_prop_of_nn,fgsm,deepfool}.

Several techniques have been proposed to generate adversarial examples. While some of these approaches~\cite{fgsm,basic_iter_method} focus on maximizing the loss function of the target model by changing the input in the opposite direction of its gradients, i.e., by using gradient ascent, others~\cite{intr_prop_of_nn,cnw_attack} modify the input using a surrogate objective function that causes the target model to misclassify the modified image. For all of these techniques, the perturbations generated are different for different images, i.e., a separate optimization has to be performed for each image to generate an adversarially perturbed image. This kind of generation of adversarial perturbations is hence called \textit{per-instance generation}.

\begin{figure}
    \includegraphics[width=\linewidth]{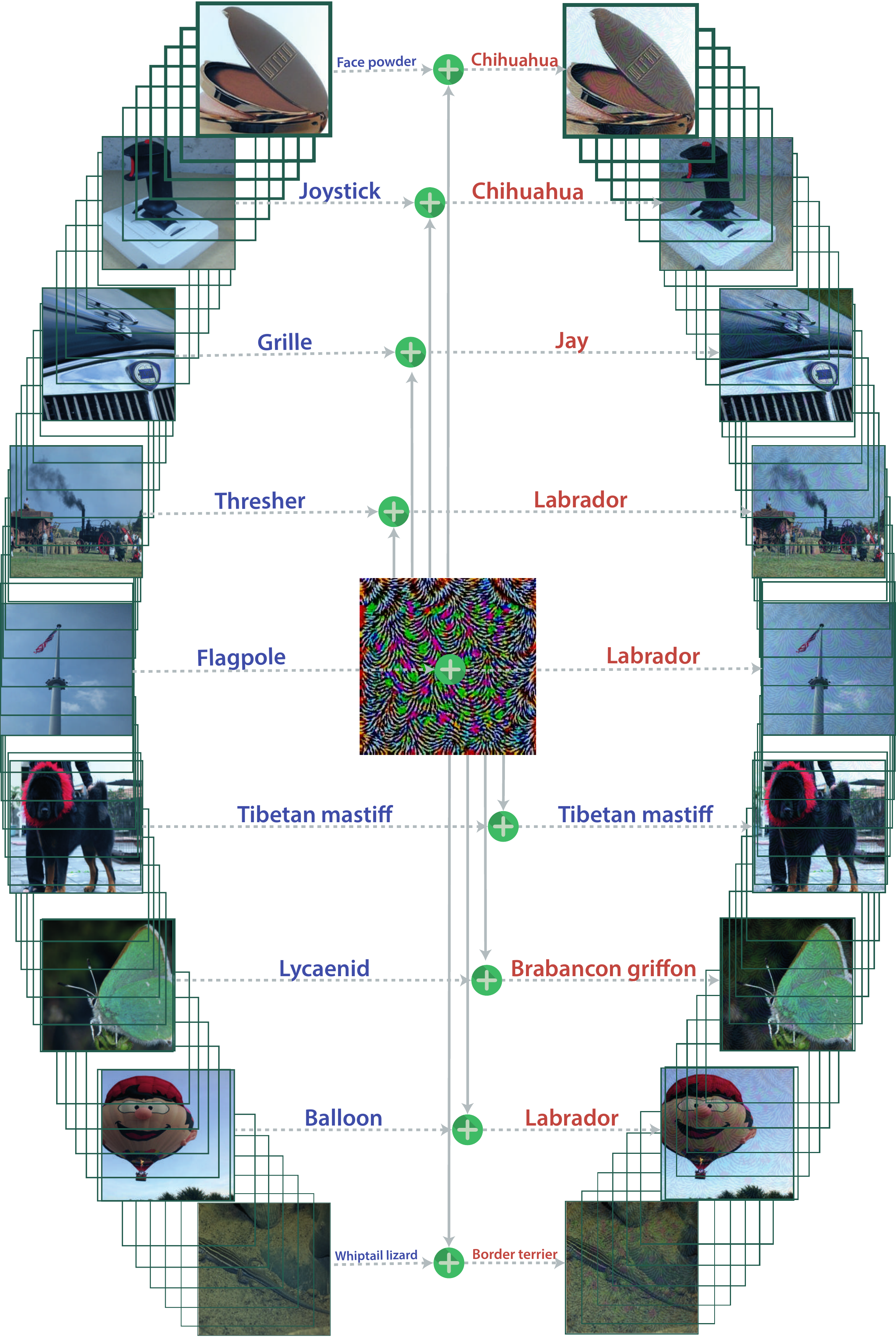}
   \caption{When the same perturbation (center) is added to clean images on the left, it causes misclassification of images generated on the right~\cite{uap_paper}. Note that the change on the clean images (left) after adding the perturbation are imperceptible.}
\label{fig:uap_main_illustration}
\end{figure}

While per-instance adversarial perturbation varies for different samples in a dataset, there exist \textit{image-agnostic perturbations} (see \autoref{subsec:terminology}) called \textit{universal adversarial perturbations (UAPs)} as introduced by Moosavi-Dezfooli \etal~\cite{uap_paper} that can fool state-of-the-art (SOTA) recognition models on most natural images with high probability and are quasi-imperceptible, i.e., not visible to the naked eye (see \textbf{\autoref{fig:uap_main_illustration}}). Since we need to compute only a single perturbation vector to fool all the images, they are much more efficient in terms of computation time and cost when compared to per-instance adversarial attacks. Generally, the $\ell_p$ norm of the perturbation is kept small to make it quasi-imperceptible. Furthermore, UAPs generalize well across different architectures, exhibiting excellent transferability and fooling rates on models other than the target model. Since~\cite{uap_paper}, many methods have been introduced by researchers to generate UAPs, both \textit{data-driven} and \textit{data-independent} (see \autoref{subsec:terminology}).~\cite{Learning_UAPs_with_GM} introduced a method to generate UAPs using Universal Adversarial Networks (UANs), leveraging Generative Adversarial Networks (GANs).~\cite{fast_feature_fool}\cite{gd_uap} introduced a data-independent approach to generate UAPs by adulterating the features extracted at multiple different layers of the network. Their approach of crafting perturbations did not utilize any knowledge about the data distribution using which the target model has been trained.~\cite{nag} introduced a data-driven approach utilizing fooling and diversity loss along with a generative model to create UAPs.~\cite{aaa_uap_class_impression} extended~\cite{nag} by introducing a two-stage process to create adversaries using class impressions without using any data. Several other works~\cite{uap_text,uap_image_retrieval,asv_uap,uap_prior_driven} have also introduced methods to create adversarial attacks.

UAPs being image-agnostic can misclassify any unseen image, making deep neural networks vulnerable to attacks. Defense mechanisms are hence required to prevent such attacks.~\cite{uap_prn} introduced the Perturbation Rectifying Network (PRN) to defend against UAPs.~\cite{uat} proposed a method to increase the robustness of a model against UAPs by training with perturbed images using min-max optimization.~\cite{uap_shared_adversarial_training} extended the previous approach~\cite{uat} by introducing a shared training procedure for defense against universal adversarial perturbation.~\cite{uap_texture_shape} analyzed the robustness of neural network models against UAPs with varying degrees of shape and texture-based training. 

\begin{figure}
    \includegraphics[width=\linewidth]{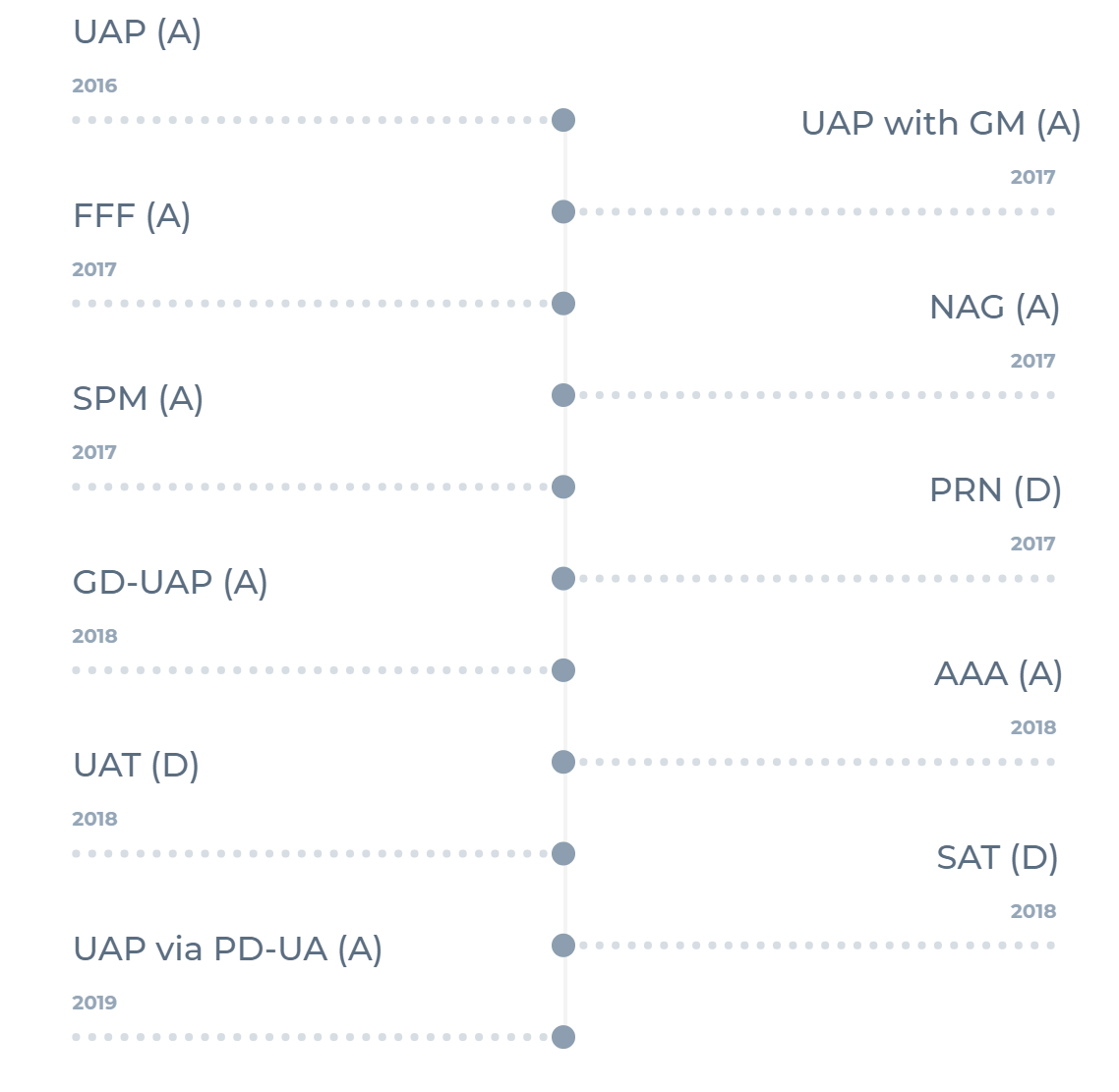}
    \caption{The above timeline shows the year of publication of various attack and defense methods mentioned in this survey. These methods include UAP~\cite{uap_paper}, UAP with GM~\cite{Learning_UAPs_with_GM}, NAG~\cite{nag}, SPM~\cite{asv_uap}, FFF~\cite{fast_feature_fool}, PRN~\cite{uap_prn}, GD-UAP~\cite{gd_uap}, AAA~\cite{aaa_uap_class_impression}, UAT~\cite{uat}, SAT~\cite{uap_shared_adversarial_training}, and UAP via PD-UA~\cite{uap_prior_driven}. (A) denotes an attacking method and (D) denotes a defense method.}
    \label{fig:my_label}
\end{figure}

Even though most of the above work were related to image classification, usage of UAPs extends to both classification and regression tasks.~\cite{gd_uap} showed the usage of UAPs in object recognition, image segmentation, and depth estimation.~\cite{uap_image_retrieval} proposed a universal adversarial attack against image retrieval.~\cite{uap_text} also showed the existence of a universal (\textit{token-agnostic}) perturbation vector that causes text to be misclassified with high probability.~\cite{uap_audio,uap_srs} showed the existence of UAPs for audio classification and speech recognition systems.

This paper aims to (i) summarize the recent advances related to universal adversarial perturbations including the various attack and defense techniques, (ii) compare and analyze various methods proposed for generating UAPs, and (iii) cover the various tasks where the applicability of UAPs has been exhibited.

The rest of the paper is organized as follows - in \autoref{sec:taxonomy}, we specify and define all the notations and terminologies used in the subsequent sections. In \autoref{sec:existence_of_UAPs}, we briefly state the reasons for the existence of universal perturbations. In \autoref{sec:attacks}, we try to cover all the different techniques of generating universal adversarial perturbations and provide a detailed comparison of these techniques. In \autoref{sec:defenses}, we cover the defense techniques that are effective against the attacks introduced in \autoref{sec:attacks}. Extension of universal perturbations to various tasks has been summarised in \autoref{sec:applications_of_uaps}. \autoref{sec:future_directions} summarizes the future directions in context to UAPs. Finally, we provide a conclusion of our survey in \autoref{sec:conclusion}.

\section{Taxonomy}
\label{sec:taxonomy}

In this section, terms and notations related to universal adversarial perturbations used throughout the paper are introduced.

\subsection{Terminology}
\label{subsec:terminology}
Following is a list of terms along with their definitions:

\begin{itemize}
    \setlength\itemsep{0.4em}
    \item \textit{image-agnostic perturbation} or \textit{universal adversarial perturbation}: Such a perturbation $\delta$ which can be added to any image $x$ to make a neural network $f$ misclassify the perturbed image $(x+\delta)$. 
    \item \textit{data-driven techniques}: Such techniques which utilize and require some images, $x \in X$, for generating adversarial perturbations.
    \item \textit{data-independent techniques}: Such techniques which do not consume any image, $x \in X$, for generating adversarial perturbations.
    \item \textit{target model}: Deep neural network under adversarial attack.
    \item \textit{white box attacks}: Attacks in which the attacker has access to the underlying training policy of the target network model.
    \item \textit{black box attacks}: Attacks in which the parameters and underlying architecture of the target network model is unknown to the attacker.
    \item \textit{non-targeted adversarial attack}: The goal of a non-targeted attack is to slightly modify the source image in a way so that it is classified incorrectly by the target model, without special preference towards any particular output.
    \item \textit{targeted adversarial attack}: The goal of a targeted attack is to slightly modify the source image in a way so that it is classified incorrectly into a specified target class by the target model.
    \item \textit{saturation rate}: The proportion of pixels $p$ out of total pixels in the perturbation $\delta$ which achieve the max-norm constraint ($\ell_\infty$) at the current iteration $t$ (used in \cite{gd_uap}).
    \item \textit{fooling rate}: The proportion of total perturbed images $(x+\delta)$ in a dataset for which $f(x) \neq f(x+\delta)$ where $f$ is the target model.
    
\end{itemize}

\subsection{Notations}

These are the mathematical notations that are followed throughout the paper. We will stick to these notations unless it is explicitly specified.

\begin{itemize}
    \setlength\itemsep{0.4em}
    \item $x$: Clean input to the target model, typically either a data sample or class impression~\cite{aaa_uap_class_impression}.
    \item $X$: Distribution of images in $\mathbb{R}^d$ using which the target model under attack has been trained, i.e., the target distribution.
    \item $X_d$: Actual data points $x_{i} \in X$ available to the attacker in case of data-driven techniques (see \autoref{subsec:terminology}).
    \item $m$: Total number of data points in $X_d$.
    \item $f$: Target model under attack, which is a trained model with frozen parameters.
    \item $K$: Total number of layers in the target network $f$.
    \item $f(x)$: Model prediction for a given data sample $x$.
    \item $f^i$: $i^{th}$ layer of the target network $f$.
    \item $f^i_k$: $k^{th}$ activation in $i^{th}$ layer of the target model.
    \item $f^{ps/m}$: Output of the pre-softmax layer.
    \item $f^{s/m}$: Output of the softmax (probability) layer.
    \item $\delta$: Additive universal adversarial perturbation (UAP).
    \item $\Delta$: Distribution of perturbations $\delta$ in $\mathbb{R}^d$.
    \item $X_\delta$: Dataset obtained by adding the perturbation $\delta$ to all the data points in $X_d$.
    \item $\xi$: Max-norm $(\ell_p)$ constraint on the UAPs, i.e., maximum allowed strength of perturbation.
    \item $S_t$: Saturation rate (see \autoref{subsec:terminology}) at the end of $t^{th}$ iteration.
    \item $F_t$: Fooling rate (see \autoref{subsec:terminology}) at the end of $t^{th}$ iteration.
    \item $H$: The patience time interval of validation for verifying the convergence of the proposed optimization (used in \cite{gd_uap}).
\end{itemize}

\section{Explaining the Existence of Universal Perturbations}
\label{sec:existence_of_UAPs}

Before diving into the various techniques of generating universal adversarial perturbations, it is essential to understand the reason behind the existence of such image-agnostic perturbations, which can be used to fool a target model $f$ on all the images in the target data distribution $X$.

\begin{figure}[h]
\centering
    \includegraphics[scale=0.5]{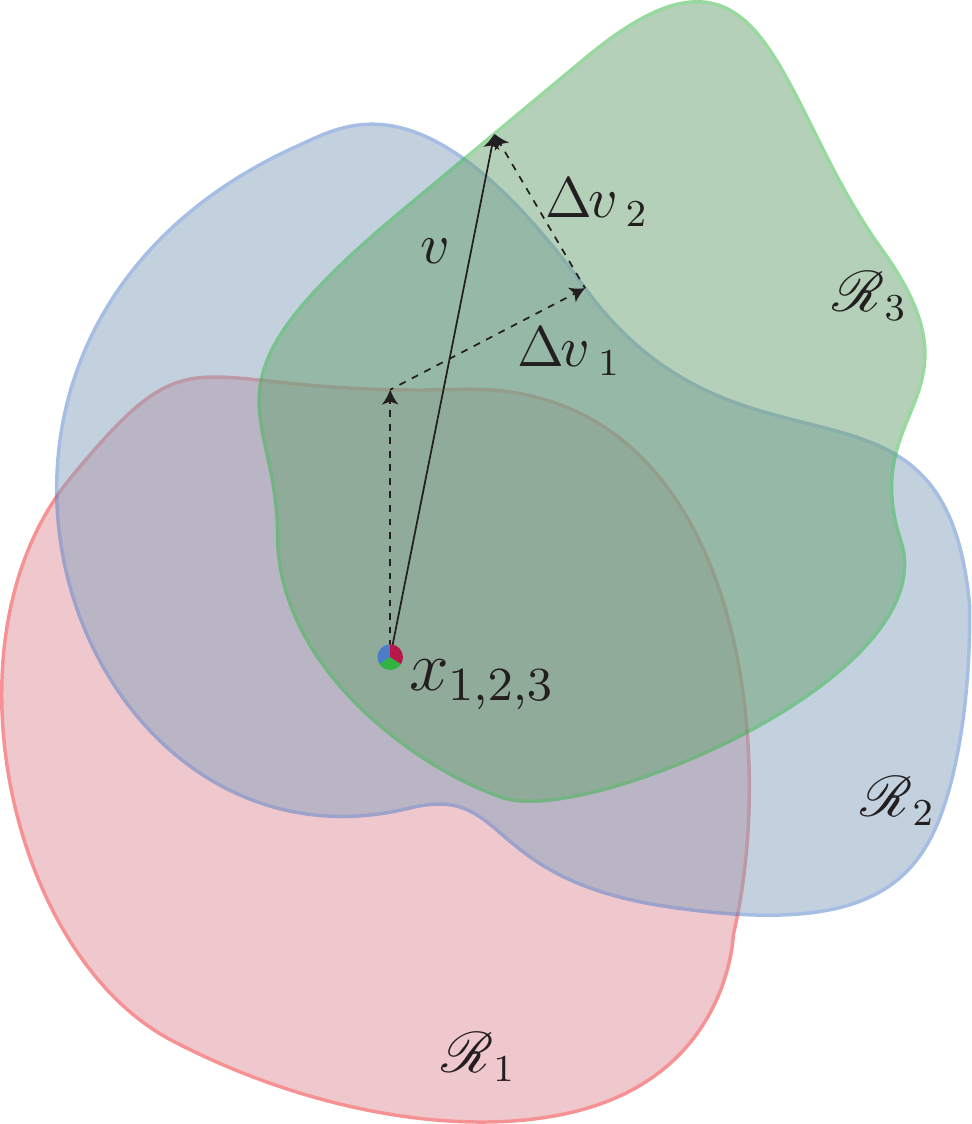}
   \caption{In this illustration, data points $x1$, $x2$ and $x3$ are super-imposed, and the classification regions $\mathscr{R}_i$ (i.e., regions of constant estimated label) are shown in different colors. One of the possible ways to get a universal perturbation~\cite{uap_paper} proceeds by aggregating sequentially the minimal perturbations sending the current perturbed points $x_i+\delta$ outside of the corresponding classification region $\mathscr{R}$.}
\label{fig:uap_explanation}
\end{figure}

For a target model $f$ to misclassify a sample data point $x \in \mathbb{R}^d$, it would be enough to add such a perturbation $\delta \in \mathbb{R}^d$ to the point $x$ so that it just crosses the decision boundary nearest to it. To get the smallest possible $\delta$, the vector $\vec{\delta}$ should be perpendicular to the nearest decision boundary. However, in the case of a universal adversarial perturbation, we want only \textit{one} such $\delta$ to be able to fool $f$ for \textit{all} the data points available in $X$. This means that our objective while generating a universal perturbation is to find such a $\delta$, which upon adding to all the data points $x$ in $X$, makes them cross the decision boundary nearest to them (see \textbf{\autoref{fig:uap_explanation}}).

At first, it might sound a challenging task to generate such a perturbation $\delta$, which simultaneously fools all the images in $X$. However, by exploiting the geometric correlations between the different parts of the decision boundary of the target model, it has been shown that generating such perturbations is possible. One of the main reasons for the existence of universal perturbations is the existence of a low dimension subspace which captures this correlation between different parts of the decision boundary of the target model $f$ \cite{uap_paper}.

\section{Attacks}
\label{sec:attacks}

Universal adversarial perturbations as introduced in \cite{uap_paper}, are such \textit{image-agnostic perturbations} $\delta$ which when added to an image $x \in X$, cause misclassification of the perturbed image $(x+\delta)$ when given as input to a target network $f$. Essentially, the main objective of a universal perturbation $\delta \in \mathbb{R}^d$ is to fool the target neural network model $f$ on almost all the input images belonging to the distribution $X$. That is, 
\begin{equation}
\label{eq:1}
    f(x+\delta) \neq f(x),\ for\ almost\ all\ x \in X
\end{equation}

In the following sub-sections, we will be covering some of the data-driven and data-independent techniques of generating universal adversarial perturbations. 

\subsection{Data-Driven Techniques}

As explained in \autoref{subsec:terminology}, data-driven techniques require some images $x \in X$ for the generation of universal perturbations. The actual number of images required for the generation of perturbations depends on the applied  approach. While some techniques~\cite{asv_uap,Learning_UAPs_with_GM} require only a small fraction of images from the target distribution $X$, others~\cite{uap_paper} require a relatively large fraction of images from the target distribution $X$.
\subsubsection{Universal Adversarial Perturbations~\cite{uap_paper}}

Along with satisfying \autoref{eq:1}, the perturbations generated by this method should also follow the max-norm constraint, i.e., the strength of the perturbations $\delta$ should not be greater than $\xi$ so that the perturbations remain imperceptible to human eyes. Also, the perturbations should achieve at least a fixed amount of fooling rate on the available data points $X_d$, which can be specified in terms of the desired accuracy ($\alpha$) on perturbed samples ($X_\delta$). These equations can be summarised as
\begin{equation}
\label{eq:2}
    \| \delta \|_p \leq \xi
\end{equation}
\begin{equation}
\label{eq:3}
    \underset{x \sim X_d}{\mathbb{P}} \left( f (x+\delta) \neq f (x) \right) \geq 1 - \alpha.
\end{equation}

\setcounter{figure}{3}

\begin{figure*}[!b]
    \centering
    \includegraphics[width=\linewidth]{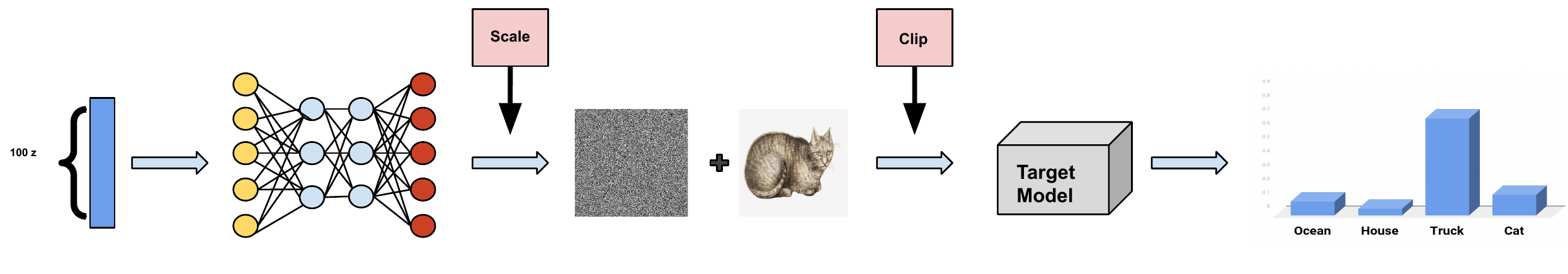}
  \caption{Overview of the model used in \cite{Learning_UAPs_with_GM}. A random sample from a normal distribution is fed into a UAN \cite{Learning_UAPs_with_GM}. This outputs a perturbation , which is then scaled and added to an image. The new image is then clipped and and fed into the target model.}
  \label{fig:gn_model}
\end{figure*}

\begin{algorithm}[h]
\caption{Iterative Deep Fool Algorithm \cite{uap_paper}}
\begin{algorithmic}[1]
\State \textbf{input:} Data points $X_d \in X$, target model $f$, desired $\ell_p$ norm of the perturbation $\xi$, desired accuracy on perturbed samples $\alpha$.
\State \textbf{output:} UAP $\delta$.
\State Initialize $\delta \gets 0^{d}$.
\While{$\text{Err} (X_\delta) \leq 1-\alpha$}
\For{each datapoint $x_i \in X_d$}
\If{$f(x_i+\delta) = f (x_i)$}
\State Compute the \textit{minimal} perturbation that sends $x_i+\delta$ to the decision boundary:
\begin{align*}
\quad \quad \Delta \delta_i \gets \arg\min_{r} \| r \|_2 \text{ s.t. } f (x_i + \delta + r) \neq f (x_i).
\end{align*}
\State Update the perturbation: 
$$\delta \gets \mathcal{P}_{p, \xi} (\delta+\Delta \delta_i).$$
\EndIf
\EndFor
\EndWhile
\end{algorithmic}
\label{alg:finding_universal_perturbations}
\end{algorithm}

To achieve the above constraints, the proposed \textbf{\autoref{alg:finding_universal_perturbations}} initializes $\delta$ with \textbf{$0^d$} and for each data point $x_i$ available in $X_d$, a smallest change $\Delta \delta_i$ is found which upon adding to $\delta$ makes the target model $f$ misclassify the perturbed data point $(x_i+\delta+\Delta\delta_i)$. The added perturbation ($\delta + \Delta \delta_i)$ is then projected on an $\ell_p$ ball of radius $\xi$ using the operation $\mathcal{P}_{p, \xi} (\delta+\Delta \delta_i)$ to get the updated $\delta$. This is done to satisfy the max-norm constraint given by \autoref{eq:2}.

To ensure constraint given by \autoref{eq:3}, all the points in the available data set $X_d$ are iterated as above until the total error $\text{Err} (X_\delta)$ achieves the desired value as specified in terms of $\alpha$. $\text{Err} (X_\delta)$ is calculated as 
\begin{equation}
\label{eq:4}
    \text{Err}(X_\delta) := \frac{1}{m} \sum_{i=1}^m 1_{f (x_i+\delta) \neq f(x_i)} \geq 1-\alpha
\end{equation}

The proposed algorithm given by \textbf{\autoref{alg:finding_universal_perturbations}}, finds one of the many possible $\delta$ satisfying \autoref{eq:2} and \autoref{eq:3} and not necessarily the optimal one. Perturbations $\delta$ generated by using the proposed algorithm \textbf{\autoref{alg:finding_universal_perturbations}} are illustrated in \textbf{\autoref{fig:uap_samples}} for various target models. We can identify some visual patterns in all of these perturbations. The perturbations for shallower networks such as VGG-16 and VGG-19~\cite{vgg_paper} have coarse patterns, while the perturbations for deeper networks such as GoogLeNet~\cite{googlenet} and ResNet-152~\cite{resnet} have much finer patterns.

\setcounter{figure}{2}

\begin{figure}[h!]
\begin{subfigure}{.25\textwidth}
  \centering
  \includegraphics[width=.8\linewidth]{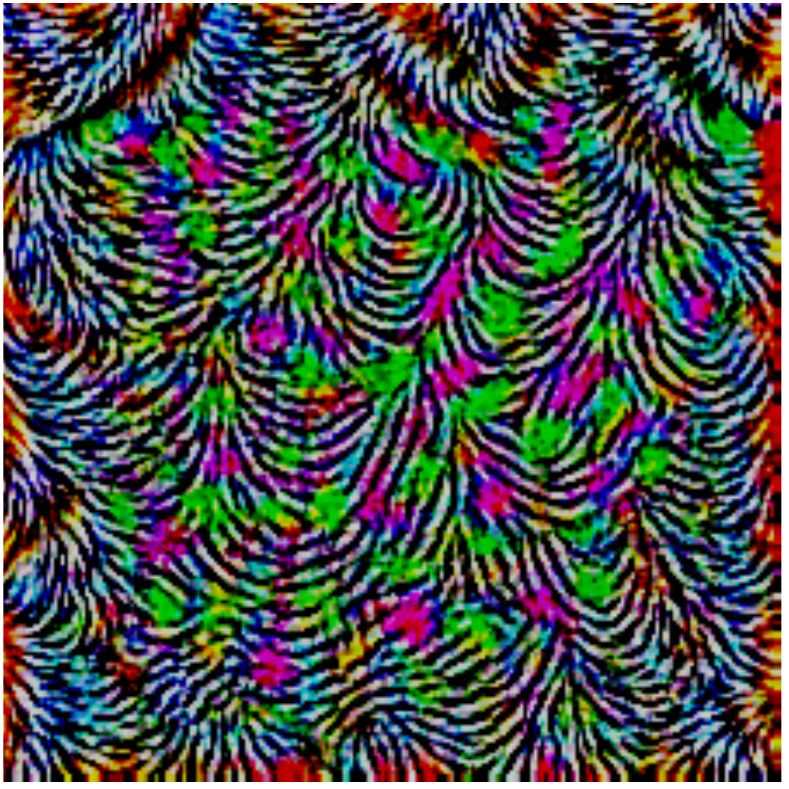}
  \caption{VGG-16}
  \label{fig:sfig1}
\end{subfigure}%
\begin{subfigure}{.25\textwidth}
  \centering
  \includegraphics[width=.8\linewidth]{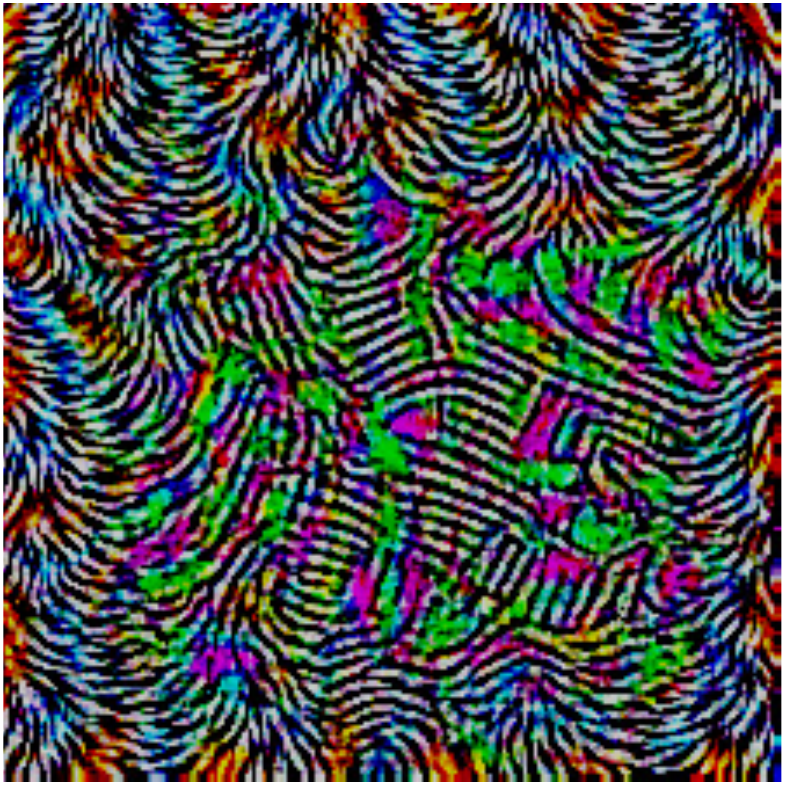}
  \caption{VGG-19}
  \label{fig:sfig2}
\end{subfigure}
\begin{subfigure}{.25\textwidth}
  \centering
  \includegraphics[width=.8\linewidth]{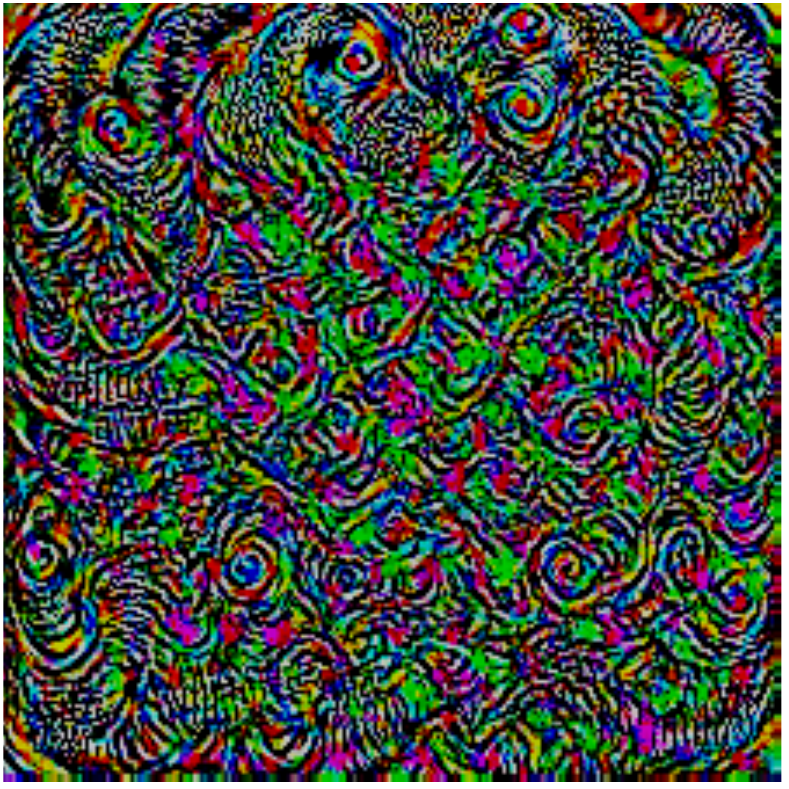}
  \caption{GoogLeNet}
  \label{fig:sfig3}
\end{subfigure}%
\begin{subfigure}{.25\textwidth}
  \centering
  \includegraphics[width=.8\linewidth]{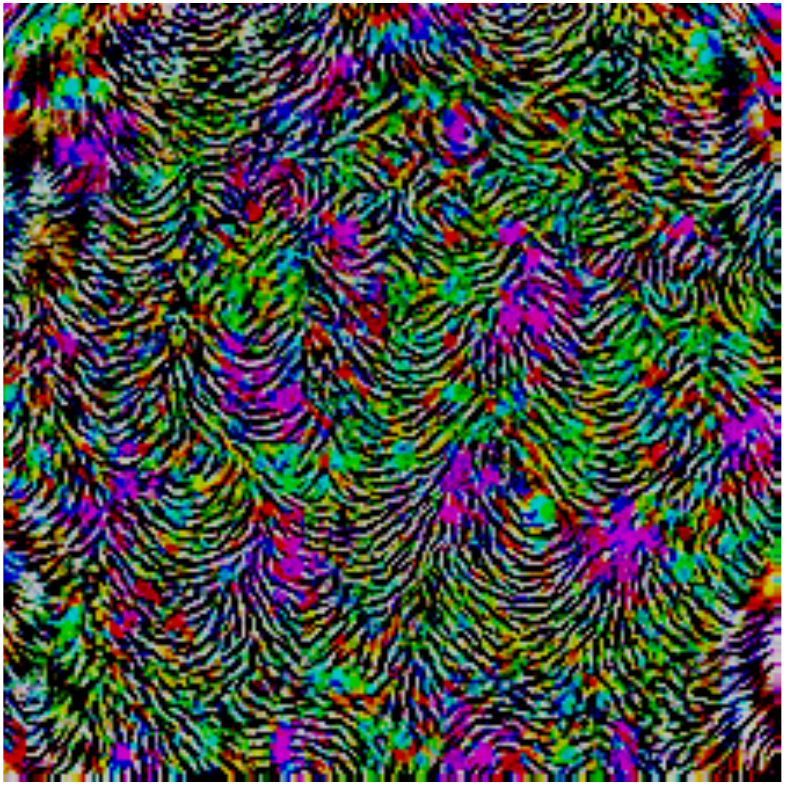}
  \caption{ResNet-152}
  \label{fig:sfig4}
\end{subfigure}
\caption{Universal perturbations generated using \autoref{alg:finding_universal_perturbations}~\cite{uap_paper} for different target models.}
\label{fig:uap_samples}
\end{figure}

\subsubsection{Learning Universal Adversarial Perturbation with Generative Models~\cite{Learning_UAPs_with_GM}}

In this attack, a generative model, namely the Universal Adversarial Network (UAN, denoted by $\mu$), is used to craft the universal perturbations. $\mu$ is trained similarly as the generator network in a GAN~\cite{gan_paper}, and hence, after training $\mu$ learns a whole distribution of perturbations rather than just a single perturbation as in \cite{uap_paper}.

An overview of the attack is illustrated in \textbf{\autoref{fig:gn_model}}. A random vector $z$ sampled from a normal distribution $\mathcal{N}(0, 1)^{100}$ is given as input to $\mu$, which outputs a perturbation $\delta$ as output. This delta is then scaled by a factor $\omega \in (0, \frac{\xi}{||\delta||_p}]$ where $\xi$ is the maximum allowed strength of the perturbation and p = (2 or $\infty$), to obtain a scaled perturbation $\delta^{'}$. The adversarial image created by adding $\delta^{'}$ to an image $x \in X_{d}$ is clipped and finally fed into the target model $f$.

\setcounter{figure}{4}

\begin{figure*}[b]
    \includegraphics[width=\textwidth]{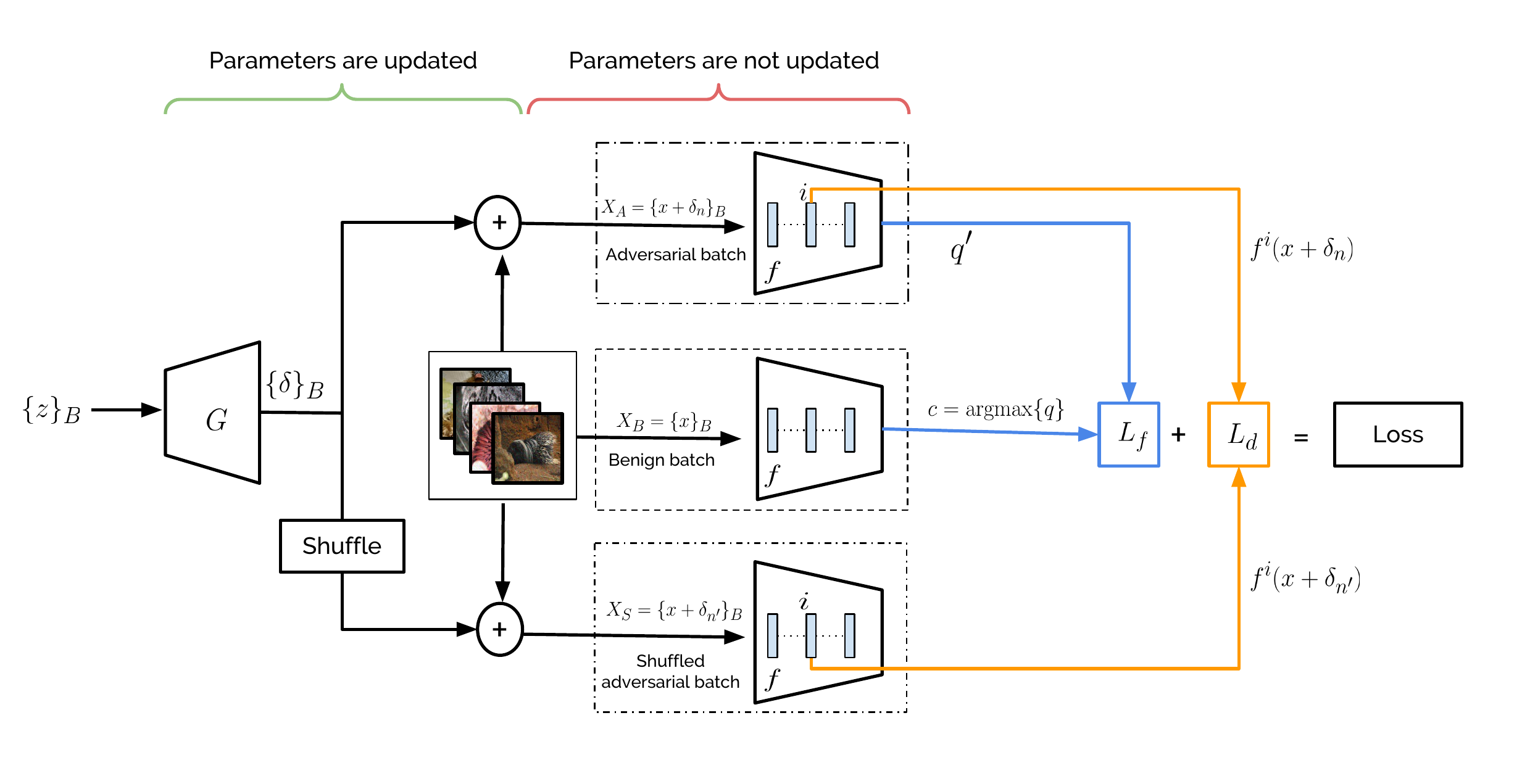}
   \caption{Overview of the approach proposed in NAG~\cite{nag} to model the distribution of generated perturbations for a given target model (Best viewed in colour).}
\label{fig:nag_fig}
\end{figure*}

The parameters of the UAN are then optimized using the following loss function
\begin{equation}
\label{eq:5}
\begin{aligned}
    L_{\text{nt}} = \max\{\log{[f(\delta^{'} + x)]_{c_0}} - \max_{i{\neq}c_0}\log{[f(\delta^{'} + x)]_{i}} - k\} \\ + \alpha\cdot||\delta^{'}||_p
\end{aligned}
\end{equation}

for a non-targeted attack where $c_0$ is the class label predicted by $f$, and
\begin{equation}
\label{eq:6}
\begin{aligned}
    L_{\text{t}} = \max\{\max_{i{\neq}c}\log{[f(\delta^{'} + x)]_{i}} - \log{[f(\delta^{'} + x)]_{c}} - k\} \\ +  \alpha\cdot||\delta^{'}||_p
\end{aligned}
\end{equation}

for a targeted attack where $c$ is the target class. It is important to note that the parameters of the target model are frozen and not optimized during the attack. 

\begin{algorithm}[h]
\caption{Learning Universal Adversarial Perturbations with Generative Models~\cite{Learning_UAPs_with_GM}}
\begin{algorithmic}[1]
\State \textbf{input:} Data points $X_d \in X$, target model $f$, desired $\ell_p$ norm of the perturbation $\xi$. 
\State \textbf{output:} Trained UAN model $\mu$ that has learned a distribution of UAPs $\Delta$.
\While{max-iteration or convergence}
\State Sample a vector z from $\mathcal{N}(0, 1)^{100}$.
\State $\delta \gets \mu(z)$.
\State $\delta^{'} \gets \omega\cdot\delta.$
\State get an image $x$ from $X_d$
\If{$\argmax_{i}f(x+\delta) = \argmax_{i}f(x)$}
    \State For non-targeted attack, optimize \autoref{eq:5}
    \State For targeted attack, optimize \autoref{eq:6}
\EndIf
\EndWhile
\end{algorithmic}
\label{alg:finding_uap_using_gm}
\end{algorithm}

Depending on the type of attack, \autoref{eq:5} or \autoref{eq:6} is optimized. The optimization is stopped once the required adversarial perturbation is found, and the confidence threshold condition is satisfied, which is given by \autoref{eq:k_gan}.

\begin{equation}
    \label{eq:k_gan}
    k > \max_{i\neq c_{0}}\log[f(\delta^{'} + x)]_{i} - \log[f(\delta^{'} + x)]_{c_{0}}
\end{equation}

Where $k$ is a hyperparameter. The complete approach can be summarized by \textbf{\autoref{alg:finding_uap_using_gm}}. The main advantage of this approach over~\cite{uap_paper} is that it allows the UAN to learn a whole distribution of universal perturbations rather than just a single perturbation. This distribution is especially useful (i) to provide an insight into the working and susceptibility of the target model, (ii) accordingly prevent black-box attacks, (iii) ease transferability of the generated perturbations across networks and increase diversity between them, as well as, (iv) facilitate adversarial training (see \autoref{sec:defenses}).

\subsubsection{NAG : Network for Adversary Generation~\cite{nag}}

Similar to the previous approach~\cite{Learning_UAPs_with_GM}, this approach also uses a generative model to learn a whole distribution $\Delta$ of UAPs. The authors have proposed novel loss functions to make sure that the generated perturbation $\delta$ successfully fools the target model, but also that these generated perturbations, are different from one another so that all the perturbations do not converge to the same perturbation.

The proposed attack is summarised in \textbf{\autoref{fig:nag_fig}}. After generation of a perturbation batch $\{\delta\}_B$, three different batches of images are formed from it to pass to the target model; namely, the \textit{adversarial batch} comprising of perturbed images$\{x+\delta\}_B$, \textit{benign batch} comprising of original images $\{x\}_B$ and the \textit{shuffled adversarial batch} which is a shuffled version of $\{x+\delta\}_B$. These batches are then passed to the target model, and the parameters of the generator model ($G$) are optimized using the following loss functions,

\begin{enumerate}
    \item Fooling loss: This loss tries to minimize the confidence of a clean label on the perturbed sample by optimizing the following \autoref{eq:fool_loss of nag} given by
    \begin{equation}\label{eq:fool_loss of nag}
    L_f = -log( 1 - f^{s/m}_{c_0})
    \end{equation}
    where $c_0$ is the class predicted by the target model for the benign batch, and $f^{s/m}_{c_0}$ is the softmax value corresponding to $c_0$ when the adversarial batch is passed through $f$.
    \item Diversity loss: This loss ensures the diversity among generated perturbations by incorporating the pair-wise distance between the generated embeddings.
    \begin{equation}\label{eq:div_loss of nag}
    L_d = - \sum_{n=1}^{B} d(f^i(x_n + \delta_n), f^i(x_n + \delta_{n}'))
    \end{equation}
    where $d(\cdot,\cdot)$ is the distance between a pair of features, $x+\delta_n$ is the $n^{th}$ perturbed image in the adversarial batch and $x_{n}+\delta_{n}^{'}$ is the $n^{th}$ perturbed image in the shuffled adversarial batch.
\end{enumerate}
The total loss function is a sum of the above two losses, 
\begin{equation}
\label{total_loss_nag}
    Total\ Loss = L_f + L_d
\end{equation}

\begin{algorithm}[!ht]
\label{alg:nag_alg}
\caption{Network for Adversary Generation~\cite{nag}}
\begin{algorithmic}[1]
\State \textbf{input}: Data points $X_d \in X$, target model $f$
\State \textbf{output}: Generator $G$ that has learned $\Delta$
\While{max iterations or convergence}
    \State Sample a vector z
    \State Transform $z$ using $G$ to get $\{\delta\}_B = (\delta_1, \delta_2, \delta_3,...,\delta_B)$
    \State Generate three quasi-batches:
\begin{itemize}
    \item Benign Batch $X_B = \{x\}_B = (x_1, x_2, x_3,...,x_B)$
    \item Adversarial Batch $X_A = (x_1 + \delta_1, x_2 + \delta_2, x_3 + \delta_3,...,x_B + \delta_B)$
    \item Shuffled Adversarial Batch $X_S = (x_1 + {\delta_1}', x_2 + {\delta_2}', x_3 + {\delta_3}',...,x_B + {\delta_B}')$ where $\delta_k$ $\neq$ $ \delta_{k}'$
\end{itemize}
\State Feed each batch through $f$ to compute loss
\State Update parameters of $G$
\EndWhile
\end{algorithmic}
\end{algorithm}

\subsubsection{Stochastic Power Method~\cite{asv_uap}}
To achieve the objective given by \autoref{eq:1}, this approach proposes to maximize the difference between the activations of a particular layer for clean image $x$ and perturbed image $x+\delta$. Mathematically,
\begin{equation}\label{eq:7}
\|f^i(x + \delta) - f^i(x) \|_q\to \max, \quad \|\delta\|_p = \xi
\end{equation}
for a small value vector $\delta$ we have,
$$f^i(x + \delta) - f^i(x) \approx J^i(x) \delta$$ where 
$$J^i(x) = \frac{\partial f^i}{\partial x} \bigg\rvert_{x}$$
is the Jacobian matrix of $f^i$.
For the $q$ norm we can write,
\begin{equation}\label{eq:8}
\|f^i(x + \delta) - f^i(x) \|_q \approx \| J^i(x) \delta \|_q
\end{equation}
To maximize the value on the L.H.S. of \autoref{eq:8}, we need to maximize the R.H.S. of \autoref{eq:8}.
Thus, the final optimization problem is reduced to 
\begin{equation}\label{eq:9}
\| J^i(x) \delta \|_q \to \max, \quad \|\delta\|_p = \xi
\end{equation}
and for the given set of data points $X_d$, \autoref{eq:9} is modified to 
\begin{equation}\label{eq:10}
\sum_{x_j \in X_d} \| J^i(x_j) \delta\|_{q} \to \max, \quad \|\delta\|_p = \xi
\end{equation}

\autoref{eq:10} can be solved by using the Power Method~\cite{power_method}. Let $X_d = \lbrace x_1, x_2 \hdots x_m \rbrace$ be the subset of the training set chosen to create the perturbations and m be the total number of samples. We compute $J^i(x_j) \in \mathbb{R}^{m \times n}$ for each $x_j \in X_d$ and stack these matrices in the following manner:
\begin{equation}\label{eq:12}
J^i(X_d) = 
\begin{bmatrix}
J^i(x_1) \\
J^i(x_2) \\
\hdots \\
J^i(x_m)
\end{bmatrix}
\end{equation}

We then compute \textit{matvec} functions of $J^i(X_d)$ and $(J^i(X_d))^T$ and run the Power Method for computing the desired $\delta$. \textbf{\autoref{alg:uap_spm}} summarizes the proposed approach (refer to the actual paper for the detailed algorithm).

\begin{algorithm}[!ht]
\caption{Stochastic Power Method for generating UAPs~\cite{asv_uap}}\label{alg:uap_spm}
\begin{algorithmic}[1]
\State \textbf{input}: A batch of images $X_d = \lbrace x_1, x_2, \hdots x_m \rbrace$, $f^i(x)$ - fixed hidden layer of the target model
\State \textbf{output}: UAP $\delta$
\For{$x_j \in X_d$}
\State Compute the Jacobian of $J^i(x_j)$ and $(J^i(x_j))^T$  
\EndFor
\State Construct the $matvec$ functions of $J^i(X_d)$ and $(J^i(X_d))^T$ defined in \autoref{eq:12}.
\State Run Power Method~\cite{power_method} with desired $p$ and $q$.
\State \textbf{return} $\delta$
\end{algorithmic}
 \label{alg:stochastic_pm}
\end{algorithm}

\subsection{Data-Independent Techniques}

As explained in \autoref{sec:taxonomy}, data-independent techniques require no data for the generation of universal adversarial perturbations. No data requirement makes these techniques more applicable in the real world, where the distribution of the target dataset is usually not available to the attacker.

\subsubsection{Fast Feature Fool~\cite{fast_feature_fool}}

The paper, for the first time, raises a reasonable concern for training data availability and thus proposes a novel data-independent approach to generate image-agnostic perturbations for a range of CNNs trained for object recognition, and shows its triple universality: (i) universality across multiple images from the target dataset over a given CNN, (ii) transferability across multiple networks trained on target dataset, and (iii) the surprising ability to fool CNNs trained on datasets different than the target dataset.

Due to the unavailability of data, the training objective has to be different from the simple flipping of network output (as used in prior data-driven approaches), so they formulated an optimization problem to compute the perturbation $\delta$ which misfires the features at individual layers of a CNN to impartially raise activations, eventually leading to the misclassification of the sample, thus fooling the CNN with high probability.

Loss function used for the computation of $\delta$ is given by,
\begin{equation}
\label{eq:fff_prod}
  Loss = - \log \left( \prod\limits_{i=1}^K \bar{f^i}(\delta) \right) \quad \text{s.t.} \quad ||\: \delta\: ||_{\infty} < \xi
\end{equation}
where, $\bar{f^i}(\delta)$ is the mean activation at layer $i$ when the perturbation $\delta$ is input to the CNN, and $K$ is the total number of layers in the CNN at which we maximize activations for the perturbation $\delta$.

The proposed objective computes the product of mean activations at multiple layers in order, and, intuitively, the product is a more definite constraint to force activations at all
layers to increase simultaneously for the loss to reduce. Also, to avoid working with extreme values ($\approx 0$), the logarithm of the product is used as the loss function. It is also worth noting that activations are considered after the non-linearity (typically ReLU). Therefore $\bar{f^i}$ is non-negative, and it is empirically suggested to restrict optimization at convolution layers.
Finally, convergence is understood when either the loss gets saturated or the fooling performance over a small held outset is maximized. The complete algorithm is summarized in \textbf{\autoref{algo:fff}}.

\begin{algorithm}[h]
\begin{algorithmic}[1]
\caption{UAP generation via FFF~\cite{fast_feature_fool}}
  \State \textbf{input:} Target model $f$, desired $\ell_\infty$ norm of the perturbation $\xi$
  \State \textbf{output:} UAP $\delta$
  \State Initialize $\delta$ randomly.
  \While{convergence}
    \State Optimize $\delta$ to achieve higher $\bar f^i(\delta)$ as in \autoref{eq:fff_prod}.  
    \State Clip $\delta$ to satisfy imperceptibility constraint $\xi$
  \EndWhile
  \State \textbf{return} $\delta$
  
\label{algo:fff}
\end{algorithmic}
\end{algorithm}

\begin{figure}
\begin{subfigure}{.25\textwidth}
  \centering
  \includegraphics[width=.8\linewidth]{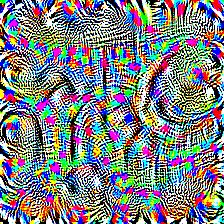}
  \caption{CaffeNet}
  \label{fig:sfig1}
\end{subfigure}%
\begin{subfigure}{.25\textwidth}
  \centering
  \includegraphics[width=.8\linewidth]{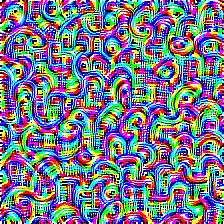}
  \caption{VGG-16}
  \label{fig:sfig2}
\end{subfigure}
\begin{subfigure}{.25\textwidth}
  \centering
  \includegraphics[width=.8\linewidth]{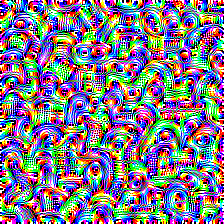}
  \caption{VGG-19}
  \label{fig:sfig3}
\end{subfigure}%
\begin{subfigure}{.25\textwidth}
  \centering
  \includegraphics[width=.8\linewidth]{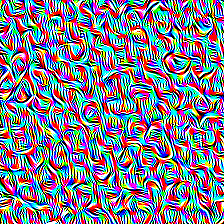}
  \caption{GoogLeNet}
  \label{fig:sfig4}
\end{subfigure}
\caption{Universal perturbations generated using Fast Feature Fool algorithm~\cite{uap_paper} for different target models.}
\label{fig:fff_uap_samples}
\end{figure}

\subsubsection{GD-UAP~\cite{gd_uap}}

This approach is an extension of the previous data-independent approach~\cite{fast_feature_fool} for generating universal adversarial perturbations. Instead of optimizing the loss given by \autoref{eq:fff_prod}, the proposed method tries to generate perturbations $\delta$ that can produce \textit{maximal spurious activations} at each layer of the target model $f$ using the following loss function,

\begin{equation}
\label{eq:gd_uap_loss_fn}
    Loss = - \log \left( \prod\limits_{i=1}^K  \Vert f^i(\delta)\Vert_2 \right) \quad \text{s.t.} \quad \Vert\:\delta\:\Vert_{\infty} < \xi
\end{equation}

Observe that the proposed loss function in \autoref{eq:gd_uap_loss_fn} uses the $\ell_2$ norm of activations of the intermediate layers instead of mean activations as used in \cite{fast_feature_fool}. The authors have also shown the effectiveness of their approach when additional priors available about the target data, such as, (i) mean ($\mu_d$) and standard deviation ($\sigma_d$) of the target data $X_d$, or (ii) the actual data points in $X_d$. 

\begin{algorithm}[htb!]
\begin{algorithmic}[1]
  \State \textbf{input:} Target model $f$, data $g$. Note that $g=0$ for data-independent case, $g= d \sim \mathcal{N}(\mu_d,\sigma_d)$ for range prior case, and $g =X_d$  for training data samples case. 
  \State \textbf{output:} Image-agnostic adversarial perturbation $\delta$
  \State Randomly initialize $\delta_0 \sim \mathcal{U}[-\xi,\: \xi]$ \\
  $t=0$\\
  $F_t=0$
  \While{$ F_t < \text{min. of } \{F_{t-H},F_{t-H+1}\ldots F_{t-1}\}$ }
    \State $t \leftarrow t+1$
    \State Compute $f^{i}(g+\delta)$ 
    \State Compute loss = $- \sum{\log \left( \prod\limits_{i=1}^K  \Vert f^i(g+\delta)\Vert_2 \right)}$ 
    \State Update $\delta_t$ through backprop.
    \State Compute the rate of saturation $S_t$ in the $\delta_t$
    \If{$ S_t < \theta  $}
     \State $\delta_t \leftarrow \delta_t/2\;$ 
    \EndIf
    \State Compute $F_t$ of $\delta_t$ on substitute dataset $D$
  \EndWhile
  \State $j \leftarrow \argmax\ \{F_{t-H},F_{t-H+1}\ldots F_{t}\}$
  \State \textbf{return} $\delta_{j}$
  \caption{GD-UAP \cite{gd_uap} using different/no priors}
\label{algo:gd_uap}
\end{algorithmic}
\end{algorithm}

Instead of scaling the generated perturbation $\delta$ at regular intervals as proposed in \cite{fast_feature_fool}, the authors have proposed adaptive rescaling of $\delta$ based on its saturation rate (see \autoref{subsec:terminology}). Rescaling of $\delta$ by half is done only when the saturation rate $S_t$ (see \autoref{sec:taxonomy}) is less than some constant $\theta$. 

The main advantage of this approach is that the authors have shown the effectiveness of their method in fooling neural networks over a wide variety of tasks such as object detection, semantic segmentation and depth estimation (see \autoref{sec:applications_of_uaps}). The complete algorithm is summarized in \textbf{\autoref{algo:gd_uap}}.

\subsubsection{Ask, Acquire and Attack: Data-free UAP Generation using Class Impressions~\cite{aaa_uap_class_impression}}

\begin{figure}[t]
    \includegraphics[width=\linewidth]{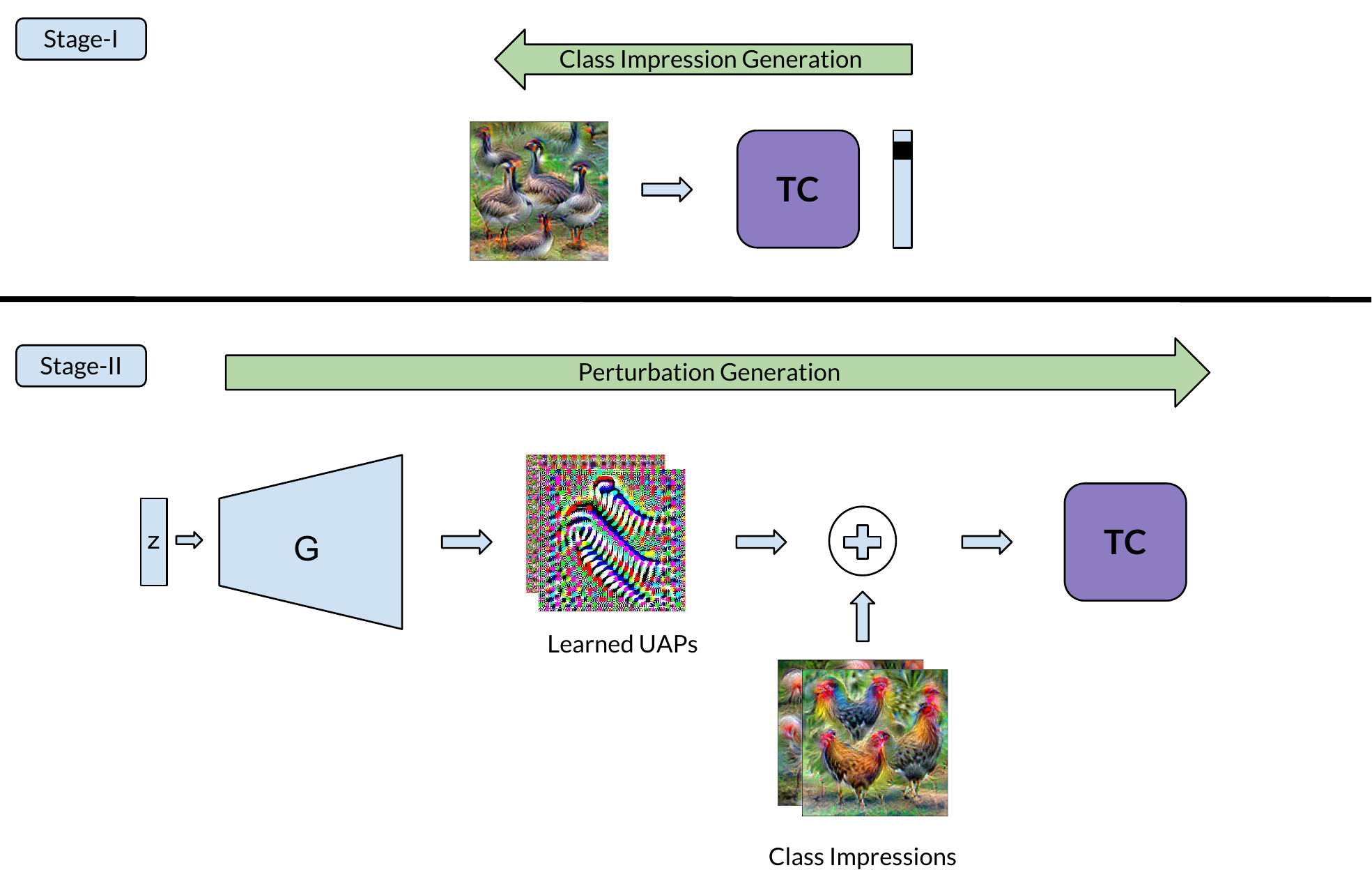}
   \caption{Overview of the proposed approach in \cite{aaa_uap_class_impression}. Stage-I, `Ask and Acquire', generates the class impressions to mimic the effect of actual data samples. Stage-II, `Attack', learns a neural network based generative model $G$ which crafts UAPs from random vectors $z$ sampled from a latent space.
}
\label{fig:aaa_uap}
\end{figure}

The data-driven approaches create UAPs by utilising the underlying data distribution and optimizing the fooling objective, thereby achieving successful fooling rates. However, data-independent approaches have access to only the parameters and architecture of the target network model, leading to lower fooling rates.  `Ask, Acquire and Attack'~\cite{aaa_uap_class_impression} is a two-stage process to craft UAPs, and tries to exploit the benefits of both data-driven and data-independent techniques (see \textbf{\autoref{fig:aaa_uap}}).
                
In the first stage, `Ask and Acquire,' the proposed methodology tries to imitate the real data samples by creating class impressions that are reconstructed from the target model's memory or trained parameters. The generation of impressions begin with a random noisy image sampled from a uniform distribution $\mathscr{U}[0, 255]$ and updating it till it changes the predicted label category with high confidence. For this purpose maximization of the predicted confidence is done in \autoref{eq:ask_acquire_loss}. Refer \autoref{sec:taxonomy} for notations used. Different random initialization results in different class impressions.

\begin{equation}
    \label{eq:ask_acquire_loss}
    CI_c = \argmax_{x} f_c^{ps/m}(x)
\end{equation}

\begin{figure}
\begin{subfigure}{.25\textwidth}
  \centering
  \includegraphics[width=.8\linewidth]{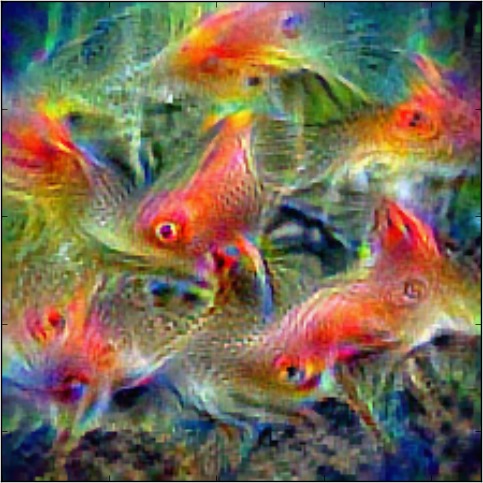}
  \caption{Goldfish}
  \label{fig:sfig1}
\end{subfigure}%
\begin{subfigure}{.25\textwidth}
  \centering
  \includegraphics[width=.8\linewidth]{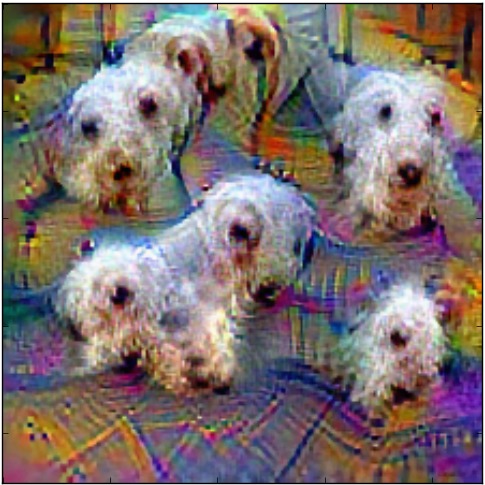}
  \caption{Lakeland terrier}
  \label{fig:sfig2}
\end{subfigure}
\begin{subfigure}{.25\textwidth}
  \centering
  \includegraphics[width=.8\linewidth]{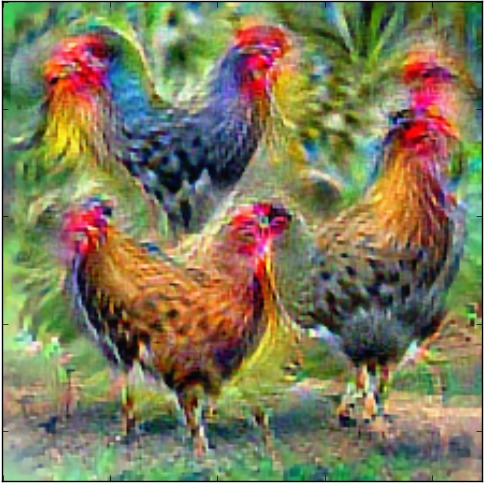}
  \caption{Cock}
  \label{fig:sfig3}
\end{subfigure}%
\begin{subfigure}{.25\textwidth}
  \centering
  \includegraphics[width=.8\linewidth]{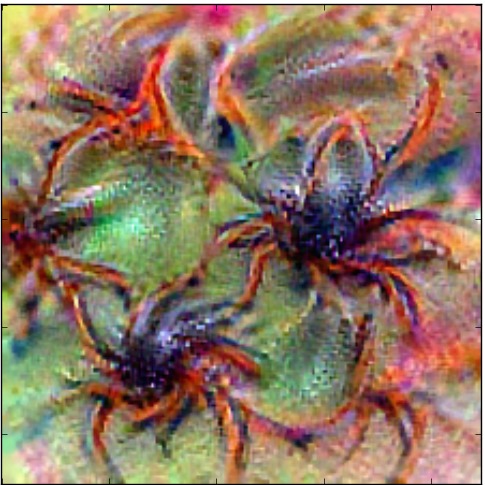}
  \caption{Wolf spider}
  \label{fig:sfig4}
\end{subfigure}
\caption{Sample class impressions generated during the `Ask and Acquire' stage of \cite{aaa_uap_class_impression} for the VGG-F target model.}
\label{fig:aaa_class_impressions}
\end{figure}

For the second stage, `Attack' a generator network $G$ is trained using the class impressions crafted in the previous stage, leveraging them as training data (see \textbf{\autoref{fig:aaa_uap}}). $G$ is trained similar to the generator of a GAN~\cite{gan_paper}. The loss functions used in this approach are inspired by~\cite{nag}. The main focus of the generator network is to generate perturbations $\delta$ which misclassify $x$ on addition. For this purpose, we try to minimize the confidence of a clean label on a perturbed sample by optimizing the fooling loss given by \autoref{eq:aaa_fooling_loss}.

\begin{equation}
    \label{eq:aaa_fooling_loss}
    L_f = -\log(1 - f_c^{s/m}(x + \delta))
\end{equation}

To ensure diversity among generated perturbations similar to \cite{nag}, the pair-wise distance between the two embeddings $f(x + \delta_i)$ and $f(x + \delta_j)$ in a mini-batch is incorporated. Thus, a diversity loss given by \autoref{eq:aaa_diversity_loss} is also proposed.

\begin{equation}
    \label{eq:aaa_diversity_loss}
    L_d = -\sum_{i,j = 1, i \neq j}^K d(f(x + \delta_i), f(x + \delta_j))
\end{equation}

Therefore total loss needed to be optimized is given by \autoref{eq:aaa_total_loss}.

\begin{equation}
    \label{eq:aaa_total_loss}
    Loss = L_f + \lambda \cdot L_d
\end{equation}

The proposed algorithm, as given by \textbf{\autoref{algo:aaa_algorithm}}, trains the generator model $G$ to create universal adversarial perturbations.

\begin{algorithm}[h]
\begin{algorithmic}[1]
\caption{Ask, Acquire and Attack: Data-free UAP Generation using Class Impressions~\cite{aaa_uap_class_impression}}
  \State \textbf{input:} Target model $f$, generator network $G$, latent space $z$, random distribution $\mu$, parameters learning rate $\lambda$, max-perturbation constraint $\xi$.
  \State \textbf{output:} Distribution of UAPs $\Delta$ learned by $G$.
  \State Create class impressions data $X_c$ for each class in the target model using \autoref{eq:ask_acquire_loss}.
  \While{max iteration or convergence}
  \For{each datapoint $x$ in $X_c$}
    \State $\delta \gets G(z)$
    \State $x^{'} \gets x + \delta $
    \State Calculate loss as mentioned in $\autoref{eq:aaa_total_loss}$.
    \State Update $G$ through backprop.
  \EndFor
  \EndWhile
\label{algo:aaa_algorithm}
\end{algorithmic}
\end{algorithm}

\subsubsection{UAP Generation via PD-UA~\cite{uap_prior_driven}}

\begin{figure}[h]
    \includegraphics[width=\linewidth]{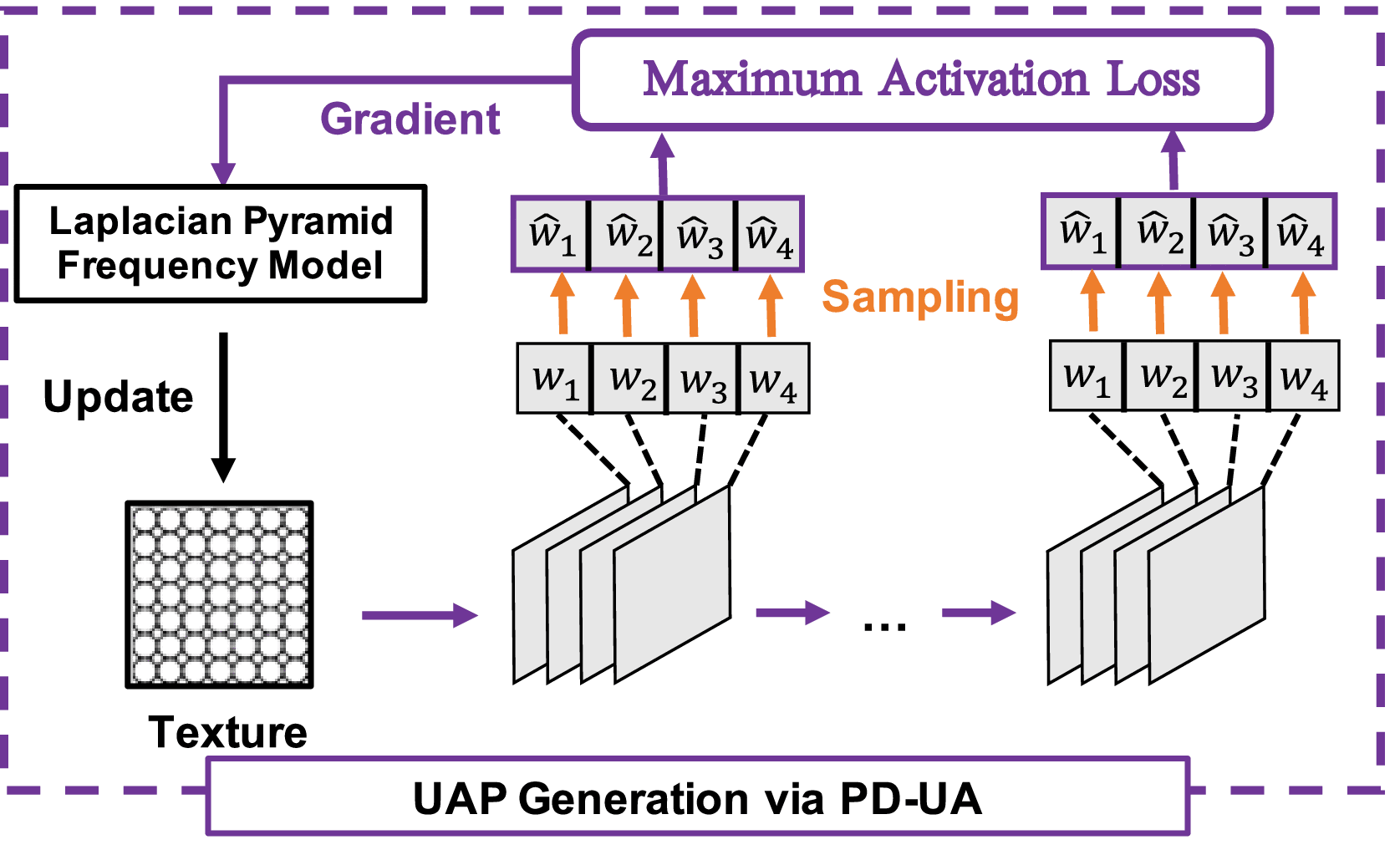}
   \caption{UAP generation technique proposed by \cite{uap_prior_driven}. Note the texture bias image at the bottom left.}
\label{fig:pd_ua_model}
\end{figure}

The proposed methodology claims that the fooling of a target model is attributed to the predictive uncertainty of the model outputs. \textit{Epistemic uncertainty} is associated with the parameters of a model that has been trained on a specific dataset (e.g., ImageNet~\cite{ILSVRC}). In contrast, the \textit{aleatoric uncertainty} is a data-independent task-dependent uncertainty which stays stable for various input data but differs for various tasks.

Proposed technique is summarised in \textbf{\autoref{fig:pd_ua_model}}. To approximate the \textit{epistemic uncertainy}, the authors have proposed the use of Monte Carlo (MC) dropout~\cite{mc_dropout} for the approximation of \textit{virtual epistemic uncertainty}~\cite{uap_prior_driven} using the following loss function for capturing the same,

\begin{equation}
\label{eq:pd_ua_epistemic}
    U_e(\delta) = \sum_{i}^{K} \sum_{k} p_e(\Delta_{ik})
\end{equation}
\begin{equation}
    p_e(\Delta_{ik}) = \frac{1}{T} \sum_t^T[-\log(\|f_j^{i}(\delta)\|_2 \cdot z_k^t)],\\
    s.t.\ \|\delta\|_p < \xi
\end{equation}

where $\Delta_{ik}$ is the $k^{th}$ neuron in the $i^{th}$ layer of the target model $f$, $z_k^t$ means the neuron $\Delta_{ik}$ is dropped out through the $t^{th}$ feedforward network, $T$ is the total number of feedforward networks for MC dropout.

\begin{algorithm}[h]
\begin{algorithmic}[1]
\caption{UAP Generation via PD-UA~\cite{uap_prior_driven}}
  \State \textbf{input:} Target model $f$, parameters learning rate $\lambda$, dropout probability $p$
  \State \textbf{output:} Image-agnostic adversarial perturbation $\delta$
  \State Initialize $\delta$ with texture image.
  \While{max iteration or convergence}
    \State Compute $f^i(\delta)$ at $i^{th}$ convolution layer. 
    \State Approximate $z_j = 0$ via MC dropout.
    \State Compute loss function as in \autoref{eq:pd_ua_loss}.
    \State Update $\delta$ through backprop.
  \EndWhile
  \State \textbf{return} $\delta_{j}$
  
\label{algo:pd_ua}
\end{algorithmic}
\end{algorithm}

A texture bias is introduced to take care of the \textit{aleatoric uncertainty}, which helps maximize the activation of neurons in the target model $f$. The Maximum Activation Loss, as given in \textbf{\autoref{fig:pd_ua_model}}, which encourages the reproduction of texture details is given as,

\begin{equation}
\label{eq:pd_ua_aleatoric}
    L_a = \mathbb{E}[\mathbb{G}_{ij}(\delta) - \mathbb{G}_{ij}(\delta_0)],\ \mathbb{G}_{ij}(\delta) = \sum_k \mathbb{F}_{ik}^l(\delta)\mathbb{F}_{jk}^l(\delta)
\end{equation}

where $\mathbb{G}$ is the Gram matrix of the features extracted from certain layers of the target model $f$, $\mathbb{F}_{ik}^l$ is the activation of the $i^{th}$ filter at position $k$ in the layer $l$, and $\delta_0$ is the texture style image that is fixed during training, as shown in \textbf{\autoref{fig:pd_ua_model}}. The combined loss function can be written as, 

\begin{equation}
\label{eq:pd_ua_loss}
    Loss = U_e(\delta) + \rho \times L_a
\end{equation}

\begin{table*}[t]
\centering
\caption{Summary of the  attributes of different attacking methods. The `Perturbation Norm' indicates the restricted $\ell_p$-norm of the perturbations that the authors of the respective methods use to show their results. The fooling rates are mentioned from the respective papers with VGG-19~\cite{vgg_paper} as the target model.}
\label{tab:1}
\begin{tabular}{|l||c|c|c|c|}
\hline
\textbf{Method}               & \textbf{Data-Driven/Independent} & \textbf{Targeted/Non-Targeted} & \textbf{Perturbation Norm} & \textbf{Fooling Rate} \\ \hline \hline
\textit{UAP}~\cite{uap_paper}                  & Data-Driven                      & Non-Targeted                   & $\ell_2, \ell_\infty$   & 77.8\%     \\
\textit{UAP with GM}~\cite{Learning_UAPs_with_GM}         & Data-Driven                      & Non-Targeted/Targeted          & $\ell_2, \ell_\infty$  & 84.6\%      \\
\textit{NAG}~\cite{nag}                  & Data-Driven                      & Non-Targeted                   & $\ell_\infty$    &   83.8\%         \\
\textit{SPM}~\cite{asv_uap}                  & Data-Driven                      & Non-Targeted                   & $\ell_\infty$   & 60.0\%            \\
\textit{FFF}~\cite{fast_feature_fool}                  & Data-Independent                 & Non-Targeted                   & $\ell_\infty$      & 43.6\%         \\
\textit{GD-UAP}~\cite{gd_uap}               & Data-Independent                 & Non-Targeted                   & $\ell_\infty$     & 40.9\%          \\
\textit{Ask, Acquire, Attack}~\cite{aaa_uap_class_impression} & Data-Independent                 & Non-Targeted                   & $\ell_1$     & 72.8\%               \\
\textit{UAP via PD-UA}~\cite{uap_prior_driven}      & Data-Independent                 & Non-Targeted                   & $\ell_\infty$      & 48.9\%          \\ \hline
\end{tabular}
\end{table*}

where $\rho$ is a tradeoff factor between the two losses. Furthermore, the authors have used a Laplacian Pyramid Frequency Model to increase the low-frequency part of the perturbation during the update step of the optimization. The overall proposed technique is given in \textbf{\autoref{algo:pd_ua}}. A detailed specification of the approach can be found in the paper~\cite{uap_prior_driven}.

\subsection{Comparison amongst Attacks}

\begin{figure}
\centering
\begin{subfigure}{.2\textwidth}
  \centering
  \includegraphics[width=.8\linewidth]{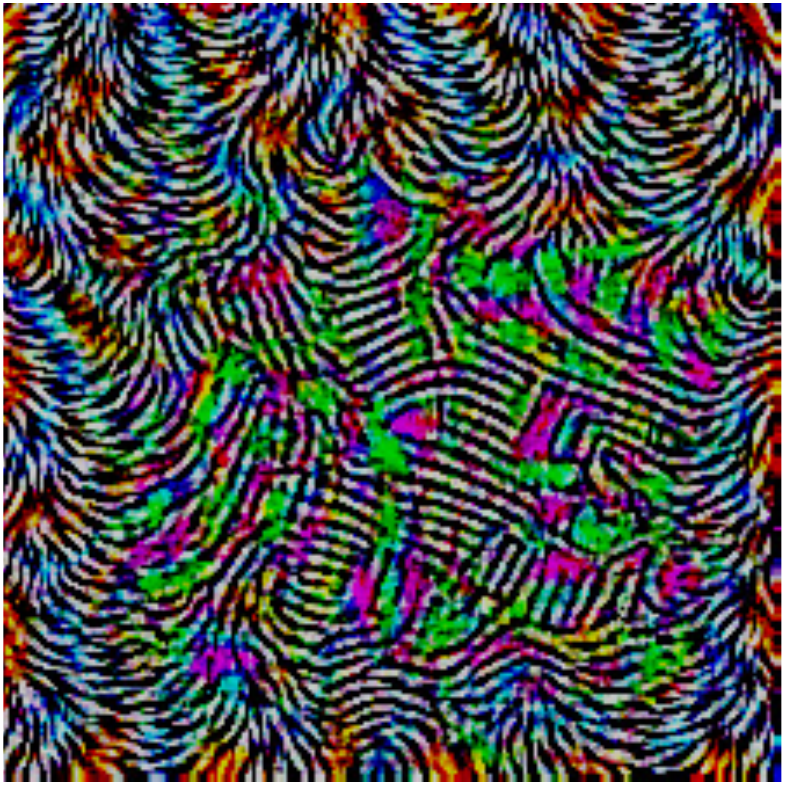}
  \caption{UAP\;\cite{uap_paper}}
  \label{fig:sfig1}
\end{subfigure}%
\begin{subfigure}{.2\textwidth}
  \centering
  \includegraphics[width=.8\linewidth]{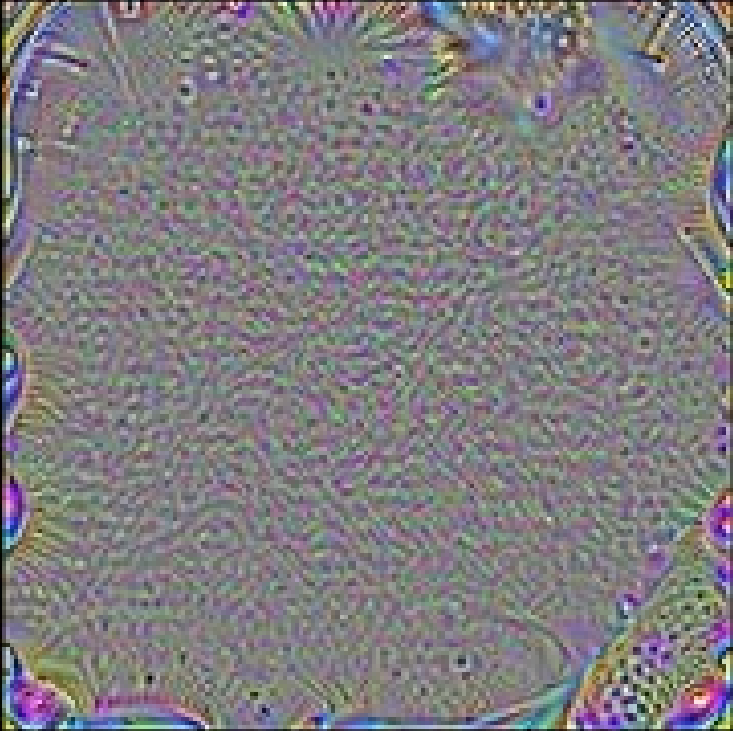}
  \caption{UAP with GM\;\cite{Learning_UAPs_with_GM}}
  \label{fig:sfig2}
\end{subfigure}
\begin{subfigure}{.2\textwidth}
  \centering
  \includegraphics[width=.8\linewidth]{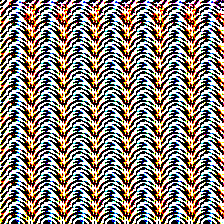}
  \caption{NAG\;\cite{nag}}
  \label{fig:sfig3}
\end{subfigure}%
\begin{subfigure}{.2\textwidth}
  \centering
  \includegraphics[width=.8\linewidth]{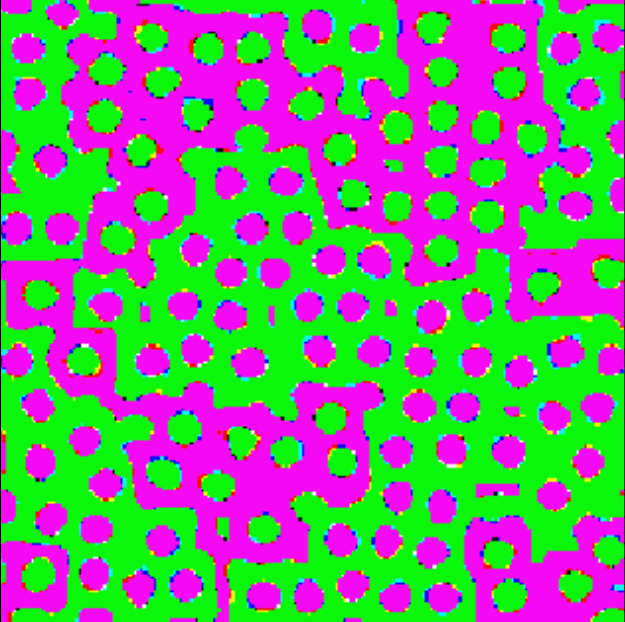}
  \caption{SPM\;\cite{asv_uap}}
  \label{fig:sfig4}
\end{subfigure}
\begin{subfigure}{.2\textwidth}
  \centering
  \includegraphics[width=.8\linewidth]{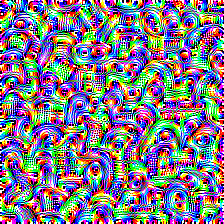}
  \caption{FFF\;\cite{fast_feature_fool}}
  \label{fig:sfig5}
\end{subfigure}%
\begin{subfigure}{.2\textwidth}
  \centering
  \includegraphics[width=.8\linewidth]{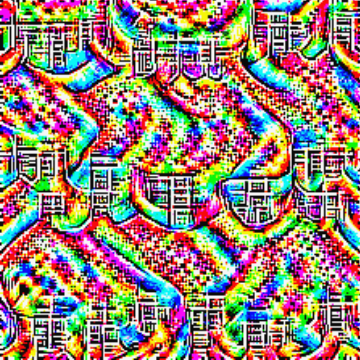}
  \caption{GD-UAP\;\cite{gd_uap}}
  \label{fig:sfig6}
\end{subfigure}
\begin{subfigure}{.2\textwidth}
  \centering
  \includegraphics[width=.8\linewidth]{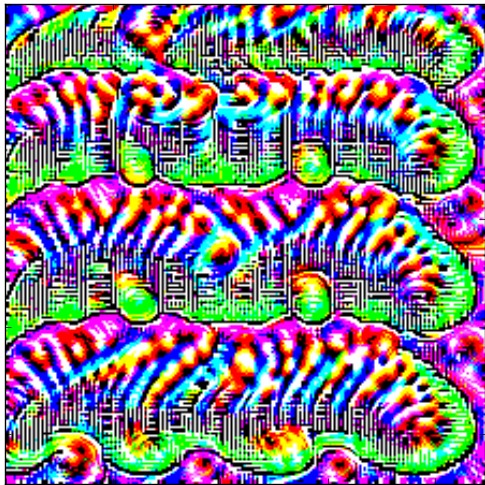}
  \caption{AAA\;\cite{aaa_uap_class_impression}}
  \label{fig:sfig7}
\end{subfigure}%
\begin{subfigure}{.2\textwidth}
  \centering
  \includegraphics[width=.8\linewidth]{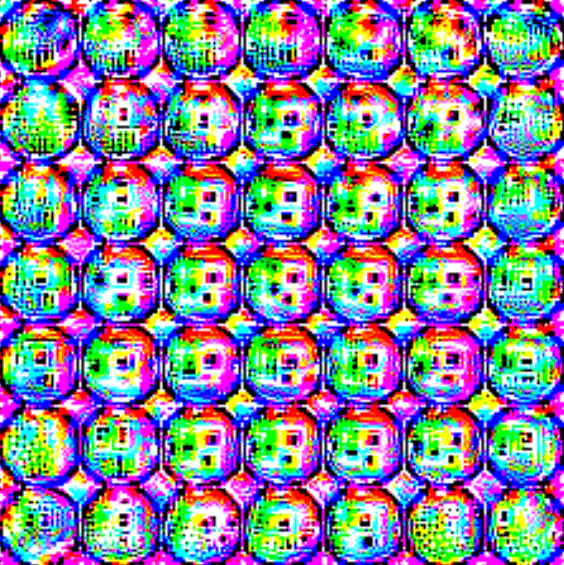}
  \caption{UAP via PD-UA\;\cite{uap_prior_driven}}
  \label{fig:sfig8}
\end{subfigure}
\caption{Universal adversarial perturbations generated by different techniques to fool the VGG-19 target model. Images are best viewed in color.}
\label{fig:aaa_class_impressions}
\end{figure}

In this sub-section, we summarize the strengths and weaknesses of the techniques covered in the previous subsections and compare them based on various parameters.

Since their introduction in \cite{uap_paper}, the techniques to produce UAPs have seen an improvement both in terms of an increase in \textit{fooling rates}, and a decrease in the number of samples in $X_d$ ($X_d = \phi$ in case of data-independent techniques) required for generating the UAP.

The fooling rates in the case of data-independent techniques \cite{aaa_uap_class_impression,uap_prior_driven} are relatively low as compared to their data-driven counterparts~\cite{nag,asv_uap}, which can be explained from the fact that the data-independent techniques do not have any prior knowledge about the distribution of the target data. 

The amount of data using which the perturbations are optimized is an important factor in determining the time of convergence of an algorithm, as all the techniques involve iterating over all the available data points in $X_d$. While \cite{uap_paper} requires a large amount of data in $X_d$ for perturbation generation, \cite{asv_uap} has shown that by using only 64 images in $X_d$ fooling rates as high as 60\% can be achieved. Since data-independent techniques do not consume any data points from the distribution $X$, these techniques converge much quicker than their data-driven counterparts but at the cost of lower fooling rates. 

Some of the techniques produce only a single universal adversarial perturbation as in \cite{uap_paper,uap_prior_driven}, while others produce a whole distribution of universal adversarial perturbations~\cite{nag,Learning_UAPs_with_GM}. Although a single universal adversarial perturbation is sufficiently capable of fooling a target model, learning a whole distribution of perturbations helps us to understand and analyze the behavior of these perturbations, and hence propose better defense techniques against them.

Although the term \textit{universal} in universal adversarial perturbation stands for the universality of the perturbation within the same data distribution $X$ using which the target model has been trained (i.e., the same perturbation can be used for all images in $X$ to fool the target network model), the perturbations produced by almost all the techniques introduced in the previous sections show excellent \textit{transferability} in fooling neural networks other than the target model~\cite{fast_feature_fool,uap_paper,nag}. Additionally, it has also been shown that the same perturbation generated for a target model can be used to misclassify images from a data distribution thats different from the target data distribution. The decrease in the fooling rate on images not in the target data distribution is significant in the case of data-driven techniques as compared to data-independent techniques~\cite{fast_feature_fool}.

\section{Defenses}
\label{sec:defenses}

Deep neural networks being vulnerable to adversarial attacks, pose a serious threat to real-world applications such as autonomous cars, biometric identification, and surveillance systems, with security being a critical factor. The vulnerability of target models towards these adversarial attacks reflects the inability of the models to learn the fundamental visual concepts. The cross-model transferability of UAPs~\cite{uap_paper,aaa_uap_class_impression} empowers attackers to generate universal perturbations rather than per-instance perturbations and also facilitates black-box attacks. Therefore, to create robust models and alleviate adversarial attacks, various defense methods were proposed, as discussed in this section.

To counter adversarial attacks, two ways of defenses are generally considered: one is to preprocess the image through various image processing algorithms at the inference time, and the other is by making the model more robust through adversarial training.

\subsection{Perturbation Rectifying Network~\cite{uap_prn}}

\begin{figure*}[t]
    \includegraphics[width=\linewidth]{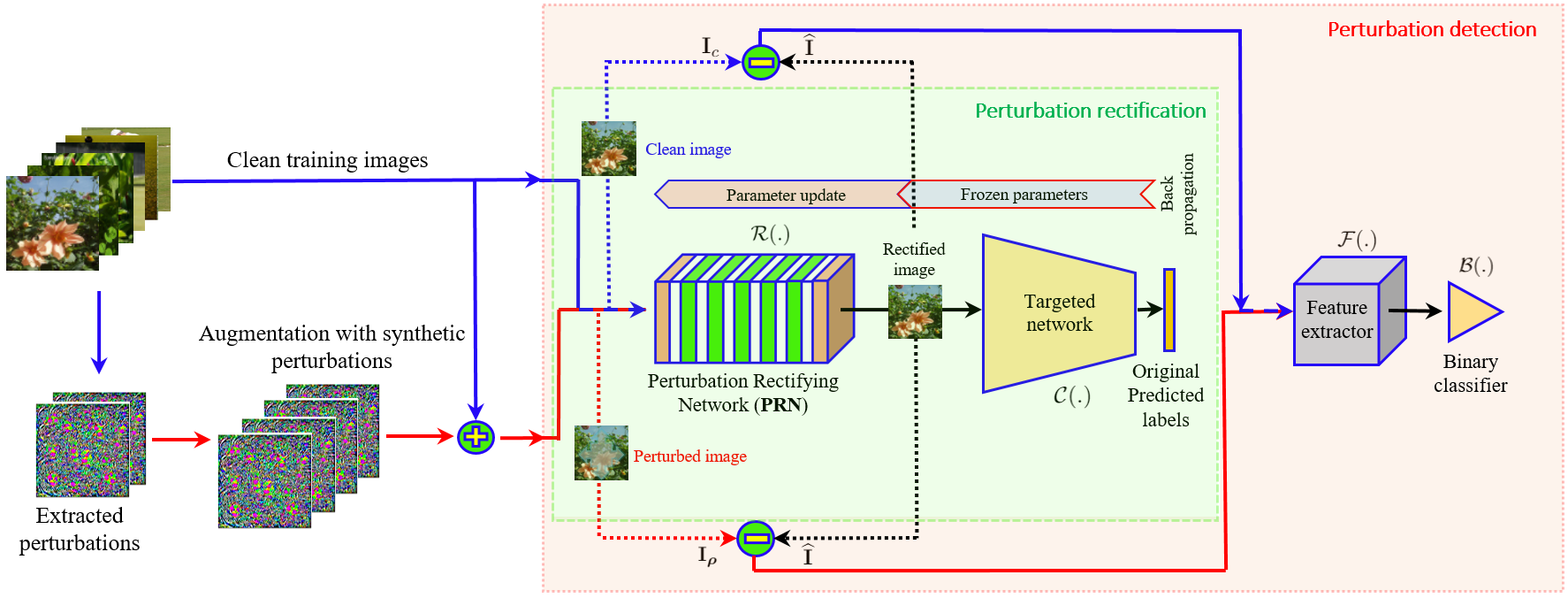}
   \caption{The process proposed in PRN \cite{uap_prn}. During the training of PRN, the loss is calculated between the label predicted by passing the clean image and the rectified image through the target model. Also, note that the weights of the target model are frozen during training.}
\label{fig:PRN_model}
\end{figure*}

This paper \cite{uap_prn} introduces a new network, namely, the \textit{Perturbation Rectifying Network (PRN)}, which can be used as a separate network to preprocess the input image before passing it to the target model. The primary purpose of the PRN is to restore the perturbed image to the original image, i.e., to subtract the perturbation from the perturbed image. So in case a perturbation is present (checked via a binary classifier called the \textit{Detector Network}), a rectified image is passed to the classifier network, else the original image is passed to the classifier network which classifies the input image. The whole process is described in \textbf{\autoref{fig:PRN_model}}. 

To train the PRN, a large amount of diverse UAPs are required to perturb clean images. However, it is very time-consuming to generate perturbations using \cite{uap_paper}. So, the authors propose an algorithm to generate synthetic perturbations using a given set of universal adversarial perturbations. 

The PRN is trained to optimize the following loss function using the set of synthetic perturbations, 
\begin{equation}
\label{eq:prn_loss}
     Loss = \frac{1}{N} \sum_{i=1}^N \mathscr{L}(l_i, l_i^*)
\end{equation}
where $\mathscr{L}(\cdot,\cdot)$ is a loss function, $l_i$ and $l_i^*$ are the labels predicted by the target model for a clean image and for the perturbed image obtained after passing through PRN. 

Also, note that the parameters of the target model which we wish to defend are frozen during the entire process. Hence, this method is compatible with defending a variety of neural network models against UAPs, and does not require the fine-tuning of the parameters of the target model.

The test time process is described in \textbf{\autoref{algo:prn}}.
\begin{algorithm}[t]
\begin{algorithmic}[1]
\caption{Method to compute correct confidence vector from a set of perturbed/non-perturbed images \cite{uap_prn}}
  \State \textbf{input:} Target model $f$, trained PRN on target model, image dataset of perturbed/non-perturbed images ($X_\delta$)
  \State \textbf{output:} Correct softmax (probability) $f^{s/m}$ of perturbed/non-perturbed images
  \While{iterate over images in $X_\delta$}
    \State Pass image($I$) through PRN and store the output as $R$. 
    \State Subtract $I$ from $R$ and pass the result to $f(.)$.
    \State Pass the output of $f(.)$ to binary classifier $B$.
    \If{$B(f(R-I))==0$ (perturbation absent)}   
     \State $f^{s/m} \leftarrow  f(I)\;$
    \ElsIf{$B(f(R-I))==1$ (perturbation present)}    
            \State $f^{s/m} \leftarrow  f(R)$
    \EndIf
  \EndWhile
\label{algo:prn}
\end{algorithmic}
\end{algorithm}

\subsection{Universal Adversarial Training~\cite{uat}}

\begin{figure*}[htb!]
    \includegraphics[width=\linewidth]{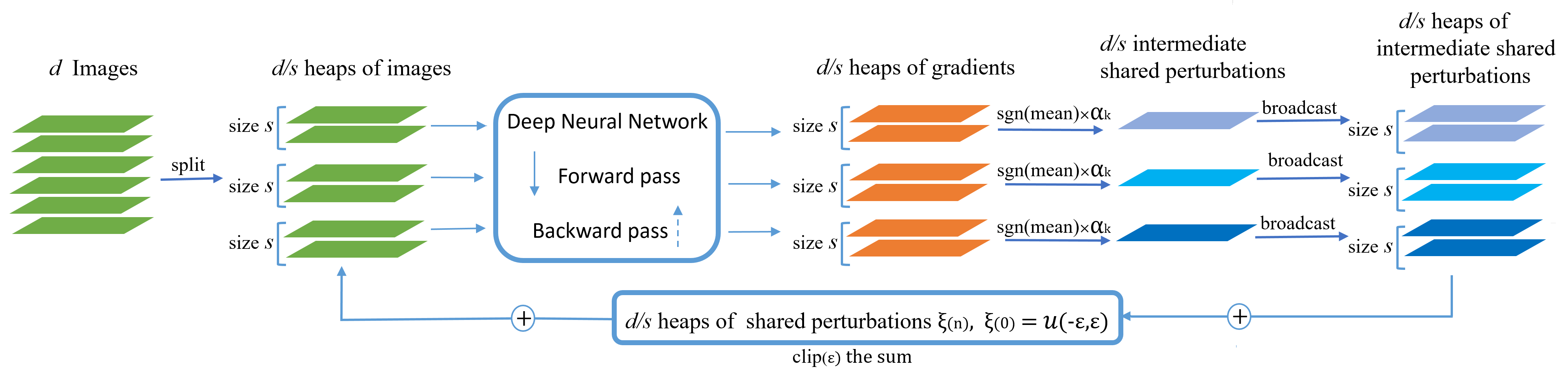}
   \caption{This figure represents the process flow as mentioned in \textbf{\autoref{alg:shared_adv_pert}}. Perturbations created through the previous step are used to train a robust model. The combined process of universal adversarial perturbation generation and training is called Shared Adversarial Training.}
\label{fig:shared_adver_model}
\end{figure*}

The paper first proposes an efficient and straightforward optimization-based universal attack that can be generated using stochastic gradient descent methods, learning perturbations 13 times faster than the standard method~\cite{uap_paper}. Then, to defend against these attacks, it proposes two methods of universal adversarial training. The first one models the training process as a two-player min-max game where the minimization is over the target model parameters, and the maximization is over the universal adversarial perturbation. The second method further improves the defense efficiency by providing a `low-cost' algorithm with a slight decrease in robustness, but reducing the training time by half as compared to the previous one.

Different from Moosavi-Dezfooli \etal~\cite{deepfool}, this paper proposed a stochastic gradient-based optimization for a $\beta$-clipped loss function,
\begin{equation}
    \max_{\delta}  \mathscr{L}(w, \delta) =  \frac{1}{N}\sum_{i=1}^{N} \hat l(w, x_i + \delta) \text{, s.t. } \| \delta \|_p \leq \xi \label{eq:univ_prob}
\end{equation}

\begin{equation}
\hat l(w, x_i + \delta) = \text{min}\{ l(w, x_i + \delta) , \, \beta\}
\end{equation}
where $l(w, \cdot)$ represents the loss used for training DNNs, $w$ represents the target model weights, $X=\{x_i; i=1,\ldots,N\}$ represents training samples, $\delta$ represents universal perturbation, and $\beta$ is hyperparameter used to clip the unbounded cross-entropy loss.

The above optimization problem can be solved by a stochastic gradient method, where each iteration is based on a minibatch of samples instead of one instance. This accelerates computation on a GPU, and requires a simple gradient update instead of the complex DeepFool~\cite{deepfool} inner loop, resulting in fast convergence of the proposed method.

\textbf{Universal adversarial training} formulates the problem of training robust classifiers as a min-max optimization problem.

\begin{equation}
    \min_{w} \max_{\|\delta\|_{p} \leq \epsilon}  (w,\delta) =  \frac{1}{N}\sum_{i=1}^{N} l(w, x_{i} + \delta)
\label{eq:adv_univ}
\end{equation}

Previously, solving this optimization problem directly
was deemed computationally infeasible due to the substantial cost
associated with generating a universal adversarial perturbation~\cite{perolat2018playing}, but the authors show that unlike Madry \etal~\cite{aleks2017deep}, updating the universal adversarial perturbation using a simple step is sufficient for building universally robust networks.
    

\begin{algorithm}
    \caption{Alternating stochastic gradient method for adversarial training against universal perturbation~\cite{uat}}
    \label{alg:univ_adv_training}
    \begin{algorithmic}[1]
        \State \textbf{input}: Training samples $X$, perturbation bound $\epsilon$, learning rate $\tau$, momentum $\mu$
        \For{epoch $= 1 \ldots N_{ep}$}
        \For{minibatch $B\subset X$}
        \State Update $w$ with momentum stochastic gradient descent;
        \State \qquad  $g_w \gets  \mu g_w  - \mathbb{}_{x \in B} [\nabla_w \, l(w, x + \delta)]$
        \State \qquad  $w \gets w + \tau g_w$ 
        \State Update $\delta$ with  stochastic gradient ascent;
        \State \qquad $\delta \gets \delta + \epsilon \, \text{sign}(\mathbb{}_{x \in B} [\nabla_{\delta} \, l(w, x + \delta)])$ 
        \State Project $\delta$ to $\ell_p$ ball.
        \EndFor
        \EndFor
    \end{algorithmic}
\end{algorithm}

As in \textbf{\autoref{alg:univ_adv_training}}, each iteration alternatively updates the target model weights $w$ using gradient descent, and then updates the universal adversarial perturbation $\delta$ using gradient ascent, only once per step, and these updates accumulate for both $w$ and $\delta$ through training.

\textbf{Low-cost Universal adversarial training:}
As UAPs are image-agnostic, so the results of the updating target model parameters and updating perturbation $\delta$ should be invariant to their order of change in one iteration. Thus the proposed methodology calls for the simultaneous update for target model parameters and the universal adversarial perturbation in \textbf{\autoref{alg:univ_adv_training}}, which backpropagates only once per iteration and produces approximately universally robust models at almost no cost in comparison to natural training.

\subsection{Defending against Universal Perturbations with Shared Adversarial Training~\cite{uap_shared_adversarial_training}}

This paper introduces the idea of `shared adversarial training’ to increase the perceptibility of universal adversarial perturbations generated through an adversarially trained model, and handle the tradeoff between enhanced robustness against UAPs vs. reduced performance on clean data samples in a better fashion than prior work.

The authors define three relevant risks:
\begin{enumerate}
    \item \textit{Expected Risk}: Expected loss of the model for the given data distribution,
    \begin{equation}
        \rho_{exp}(\theta) = \mathbb{E}_{(x, y) \sim \mathcal{D}} \, L(\theta, x, y)
    \end{equation}
    \item \textit{Adversarial Risk}: Expected adversarial loss dependent on specific samples,
    \begin{equation}
        \rho_{adv}(\theta, \mathcal{S}) = \mathbb{E}_{(x, y) \sim \mathcal{D}} \left[\sup\limits_{\xi(x) \in \mathcal{S}} L(\theta, x + \xi(x), y) \right]
    \end{equation}
    \item \textit{Universal Adversarial Risk}: Expected Adversarial Loss generalised over the entire data distribution,
    \begin{equation}
        \rho_{uni}(\theta, \mathcal{S}) = \sup\limits_{\xi \in \mathcal{S}} \mathbb{E}_{(x, y) \sim \mathcal{D}} \left[L(\theta, x + \xi, y) \right]
    \end{equation}
\end{enumerate}
And relate them as, 
\begin{equation}
    \rho_{exp}(\theta) \leq \rho_{uni}(\theta, \mathcal{S}) \leq \rho_{adv}(\theta, \mathcal{S}), \;\; \forall \theta \;\; \forall \mathcal{S} \supset \{\mathbf{0}\}
\end{equation}

Here the set $\mathcal{S}$ defines the space from which perturbations may
be chosen. The objective of adversarial training is to minimise $\sigma \cdot \rho_{adv}(\theta, \mathcal{S}) + (1 - \sigma) \cdot \rho_{exp}(\theta)$. For this work, an adversary has been defined as a function that maps data points and a set of model parameters to find a strong perturbation that maximizes the fooling rate. A special kind of adversary called heap adversary is further introduced to compute perturbations against each of the heaps generated, as shown in \textbf{\autoref{fig:shared_adver_model}}. For further clarification regarding risks, refer~\cite{uap_shared_adversarial_training}. The proposed algorithm given by \textbf{\autoref{alg:shared_adv_pert}} finds $\delta$ through shared adversarial training.

\begin{algorithm}[!ht]
\caption{Method to generate shared adversarial perturbations~\cite{uap_shared_adversarial_training}}
\label{alg:shared_adv_pert}
\begin{algorithmic}[1]
\State \textbf{input}: Data points $d$, sharedness $s$
\State \textbf{output}: $\delta$ to be used for adversarial training
\State Initialize $delta$.
\For{ $d/s$ iterations}
\State Generate gradient $\nabla$ by passing through target model
\State $\nabla$ $\leftarrow$ $sgn(\nabla)*\alpha_k$
\State Broadcast $\nabla$ to size $s$
\State $\delta$ $\leftarrow$ $\delta$ $+$ generated shared perturbation $\nabla$
\State clip $\delta$ to $[-\xi, \xi]$
\EndFor
\State Add shared perturbations $\delta$ to $d/s$ heaps for adversarial training. 
\State Iterate till convergence.
\State \textbf{return} $\delta$
\end{algorithmic}
\end{algorithm}

The benefits of shared adversarial training in this paper are proved to be extended to image segmentation tasks, apart from the usual image classification tasks. From the experiments, it is also noted that this technique leads to models that are particularly robust against universal adversarial perturbations.

\section{Extension of UAPs to Other Tasks}
\label{sec:applications_of_uaps}

In all of the previous sections, the focus was mainly on the techniques which generate UAPs to fool the target model trained for classification tasks. In this section, we cover all the various tasks where UAPs can be effectively used.

\subsection{Semantic Segmentation}

\begin{figure*}[t]
    \centering
    \includegraphics[width=\linewidth]{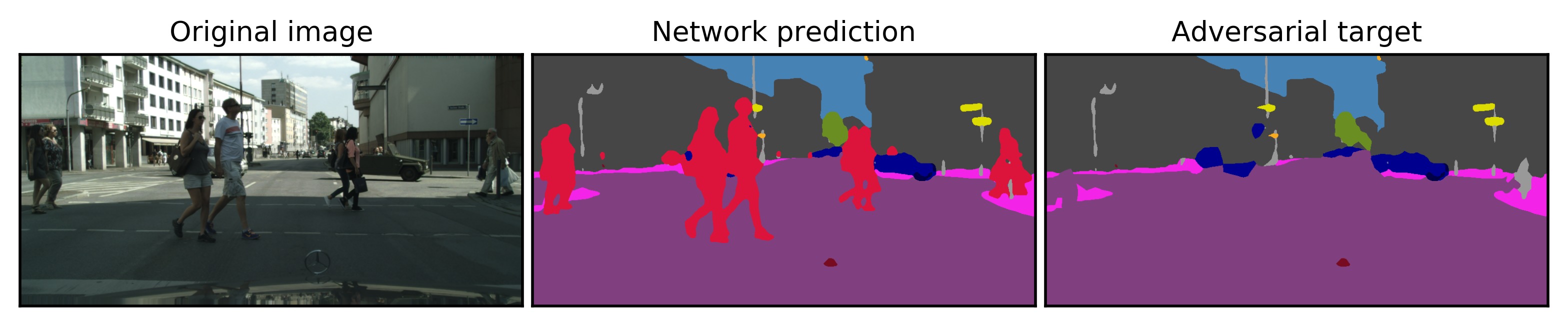}
    \caption{Illustration of the effect of adding perturbation in generating a target segmentation map to hide pedestrians.}
    \label{fig:segmen_pedestrians_uap}
\end{figure*}

Semantic segmentation is the task of assigning a class to every pixel in the image. Various applications rely on semantic segmentation, whether it is route navigation for autonomous cars, robot vision, or portrait mode of a smartphone. Before diving into how adversarial attacks for semantic segmentation are generated, it is important to exemplify the fooling in semantic segmentation. There exist adversarial perturbations that cause the model to output the same segmentation map for different arbitrary inputs. The attacker could also target removing a specific class, leaving the rest of the segmentation map unchanged~\cite{semantic_uap_other}. As seen in \textbf{\autoref{fig:segmen_pedestrians_uap}}, adding the perturbation hides a specific class (pedestrian in this case) from the target segmentation map. This type of attack poses a severe threat to the real-world application of semantic segmentation in autonomous cars, robotics, or other computer vision-related tasks. Adversarial attacks for segmentation tasks are quite similar to the classification task. While the fooling rate for an image classification task is well defined, it is unclear for other tasks like segmentation. Therefore a task-independent `Generalized Fooling Rate (GFR)' is introduced here,

\begin{equation}
    \label{eq:GFR}
    GFR(M) = \frac{R - M(\hat{y}_{\delta}, \hat{y})}{R}
\end{equation}

Here $M$ is the metric for measuring the performance of a model for any given task, $\hat{y}$ is the clean image output, $\hat{y}_{\delta}$ is the perturbed image output, and $R$ is the range of $M$.

\begin{figure}[b]
\centering
\noindent\begin{minipage}{\linewidth}
  \centering
  \begin{minipage}{.22\textwidth}
      \centering
    \includegraphics[width=\linewidth]{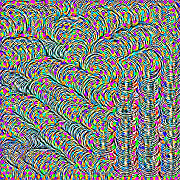}\
    FCN-Alex
  \end{minipage}
   \begin{minipage}{.22\textwidth}
       \centering
    \includegraphics[width=\linewidth]{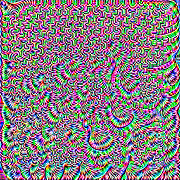}\
    {\small FCN-8s-VGG}
  \end{minipage}
  \begin{minipage}{.22\textwidth}
   \centering
    \includegraphics[width=\linewidth]{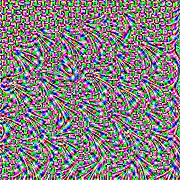}\
    DL-VGG
  \end{minipage}
  \begin{minipage}{.22\textwidth}
      \centering
    \includegraphics[width=\linewidth]{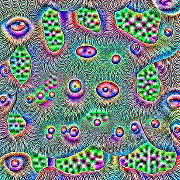}\
    DL-RN101
  \end{minipage}
    \vspace{0.002\textwidth}
\end{minipage}
\vspace{.01cm}
\caption{UAPs crafted by GD-UAP (see \autoref{algo:gd_uap}) for semantic segmentation with initial prior as 'data with less BG.'}
\label{fig:data-free-sample-perturbations-segmentation}
\end{figure}

\begin{figure}[t]
\centering
\noindent\begin{minipage}{\columnwidth}
  \centering
  \begin{minipage}{.20\linewidth}
      \centering
    \includegraphics[width=\textwidth]{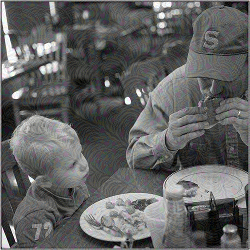}\
  \end{minipage}
   \begin{minipage}{.20\textwidth}
       \centering
    \includegraphics[width=\textwidth]{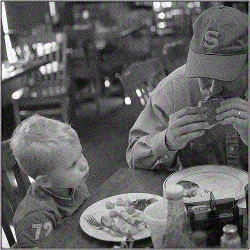}\
  \end{minipage}
  \begin{minipage}{.20\textwidth}
      \centering
    \includegraphics[width=\textwidth]{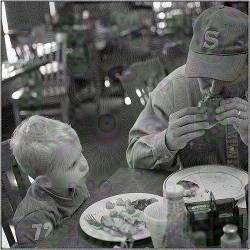}\
  \end{minipage}
  \begin{minipage}{.20\textwidth}
      \centering
    \includegraphics[width=\textwidth]{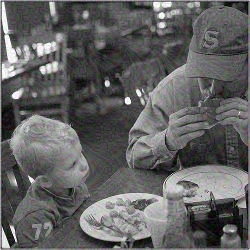}\
  \end{minipage}
  \vspace{0.002\textwidth}
\end{minipage}
\noindent\begin{minipage}{\columnwidth}
  \centering
  \begin{minipage}{.20\textwidth}
      \centering
    \includegraphics[width=\textwidth]{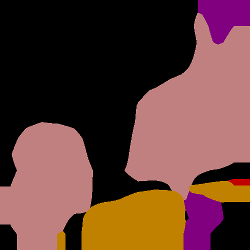}\
  \end{minipage}
   \begin{minipage}{.20\textwidth}
       \centering
    \includegraphics[width=\textwidth]{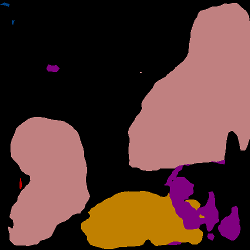}\
  \end{minipage}
  \begin{minipage}{.20\textwidth}
      \centering
    \includegraphics[width=\textwidth]{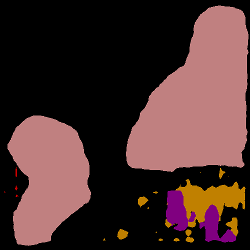}\
  \end{minipage}
  \begin{minipage}{.20\textwidth}
      \centering
    \includegraphics[width=\textwidth]{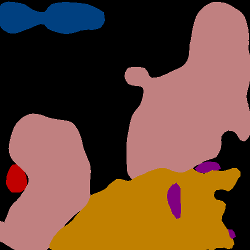}\
  \end{minipage}
  \vspace{0.004\textwidth}
\end{minipage}
\noindent\begin{minipage}{\columnwidth}
  \centering
  \begin{minipage}{.20\textwidth}
  \centering
    \includegraphics[width=\textwidth]{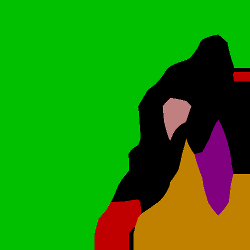}\
    FCN\\Alex
  \end{minipage}
   \begin{minipage}{.2\textwidth}
   \centering
    \includegraphics[width=\textwidth]{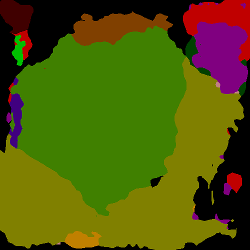}\
    FCN-8S\\VGG
  \end{minipage}
  \begin{minipage}{.2\textwidth}
      \centering
    \includegraphics[width=\textwidth]{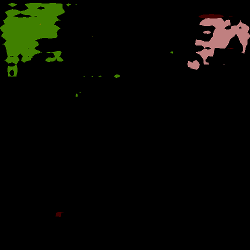}\
    DL\\RN101
  \end{minipage}
  \begin{minipage}{.2\textwidth}
  \centering
    \includegraphics[width=\textwidth]{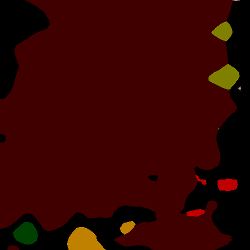}\
    DL\\VGG
  \end{minipage}
    \vspace{0.002\textwidth}
\end{minipage}
\vspace{0.1cm}
\caption{Semantic segmentation output with initial prior as 'data with less BG' for different models. The first row shows perturbed images, the second row shows segmentation output for clean images, and the third row shows segmentation output for perturbed images.}
\label{fig:multiple-nets-segmentation}
\end{figure}

\begin{figure*}[t]
\centering
\noindent\begin{minipage}{\textwidth}
  \centering
  \begin{minipage}{.23\textwidth}
      \centering
    \includegraphics[width=\linewidth]{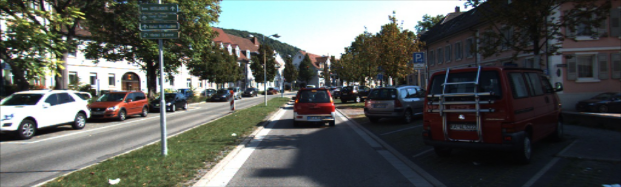}\
  \end{minipage}
   \begin{minipage}{.23\textwidth}
       \centering
    \includegraphics[width=\linewidth]{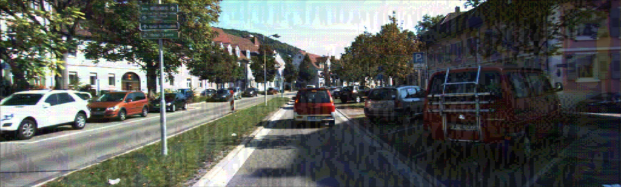}\
  \end{minipage}
  \begin{minipage}{.23\textwidth}
      \centering
    \includegraphics[width=\linewidth]{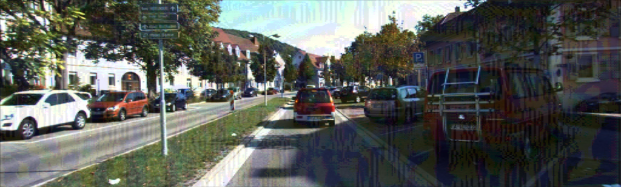}\
  \end{minipage}
  \begin{minipage}{.23\textwidth}
      \centering
    \includegraphics[width=\linewidth]{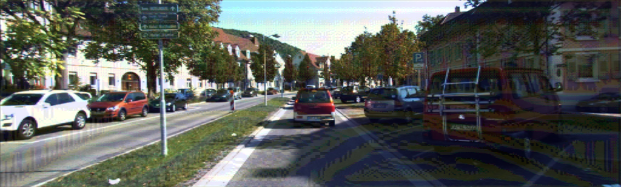}\
  \end{minipage}
  \vspace{0.002\textwidth}
\end{minipage}
\noindent\begin{minipage}{\textwidth}
  \centering
  \begin{minipage}{.23\textwidth}
      \centering
    \includegraphics[width=\linewidth]{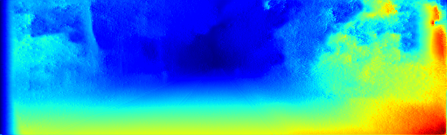}\
  \end{minipage}
   \begin{minipage}{.23\textwidth}
       \centering
    \includegraphics[width=\linewidth]{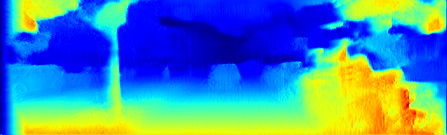}\
  \end{minipage}
  \begin{minipage}{.23\textwidth}
      \centering
    \includegraphics[width=\linewidth]{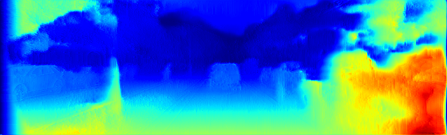}\
  \end{minipage}
  \begin{minipage}{.23\textwidth}
      \centering
    \includegraphics[width=\linewidth]{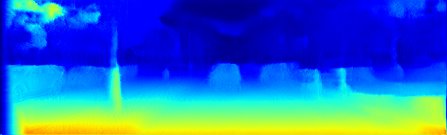}\
  \end{minipage}
  \vspace{0.002\textwidth}
\end{minipage}
\vspace{0.1cm}
\caption{The first row shows clean and perturbed images with various priors, and the second row shows the corresponding predicted depth maps from the KITTI~\cite{Geiger2012CVPR} dataset generated for monodepth-VGG.}
\label{fig:sample-depth-images-vgg}
\end{figure*}

Defining the fooling rate in such a manner helps in effectively measuring the changes caused by the adversaries in the model's output. It also assists the attacker in determining the performance of the perturbation in terms of the damage caused to a model with respect to a metric.

Semantic segmentation models are trained to perform pixel-level classification into $n$ categories, including the background. Performance is generally measured through mean Intersection over Union (IoU) computed between the predicted and ground truth segmentation map. Extension of UAPs~\cite{uap_paper} directly for this task is non-trivial.~\cite{gd_uap} showed that their algorithm for the generation of UAPs could be directly applied for this task without any changes. The authors reported their results on the PASCAL-VOC 2012~\cite{pascal-voc-2012} dataset. Since this dataset had the background pixels percentage greater than half, the authors created a smaller training dataset with a lower background pixels percentage and named it as 'data with less BG'. The perturbation created with this data has a higher capability of fooling or an increased Generalized Fooling Rate for mean IoU (GFR(mIoU)).

As seen in \textbf{\autoref{fig:data-free-sample-perturbations-segmentation}}, perturbations learned for the task of semantic segmentation for different models look different across architectures, similar to object recognition. For the same input image, perturbations learned by different models produce very different outputs even when their clean image output looks similar, as can be noted from \textbf{\autoref{fig:multiple-nets-segmentation}}.

Metzen \etal\cite{hendrik2017universal} proposed a novel approach to generate UAPs specifically for semantic segmentation. They introduced two methods, one to make the target model output a fixed segmentation map, and the other to remove one class from the target model output. They further show that on passing the generated UAPs (in case of a fixed target segmentation map) through the target model, the output looks similar to the target scene itself.

~\cite{uap_shared_adversarial_training} proposed shared adversarial training for training robust models against universal adversarial perturbations for the semantic segmentation task. Their method showed a considerable increase in robustness for both targeted and non-targeted attacks for a little tradeoff in accuracy.

\subsection{Depth Estimation}

Most of the attacks discussed until now mainly focused on creating UAPs for classification tasks. Various recent works such as \cite{uap_regression_1,uap_regression_2,uap_regression_3} showed an increase in the use of convolutional networks for computer-based regression tasks. Depth estimation is the task of obtaining a depth map of an RGB image. For monocular-depth estimation, it can also be framed as a \textit{pixel-level continuous regression} problem. Algorithms to calculate the depth map do not rely on hand-crafted features and instead use deep neural networks. Adversarial attacks on these models may cause the network to regress the depth per pixel erroneously. Targeted attacks on these networks result in the depth of a specific class to be inaccurate (see \textbf{\autoref{fig:sample-depth-images-vgg}}).~\cite{gd_uap} provided an algorithm for crafting UAPs for CNNs performing regression. Similar to semantic segmentation, the fooling rate for depth estimation is reported through Generalized Fooling Rate (see\autoref{eq:GFR}). The authors showed that the fooling performance of the perturbation also varies with the metric used for analysis, hence reported fooling through $GFR(\delta)$, $GFR(Abs.Rel.Err.)$ and $GFR(RMSE)$. 

Universal adversarial perturbations generally target non-robust features of an image for fooling.~\cite{depth_estimation_uap} conjectured that \textit{"the attacks are made possible not by perturbing salient pixels containing important depth cues, but by mostly perturbing non-salient pixels."}. Based on this conjecture, they introduced a training regime for creating a robust model, proposing to mask out the non-salient pixels using a saliency map\footnote{\textit{Saliency Map: a set of a small number of pixels from which a CNN can estimate depth accurately.}} since they are the most vulnerable to adversarial attacks.

\subsection{Image Retrieval}

\begin{figure}[h]
    \includegraphics[width=\linewidth]{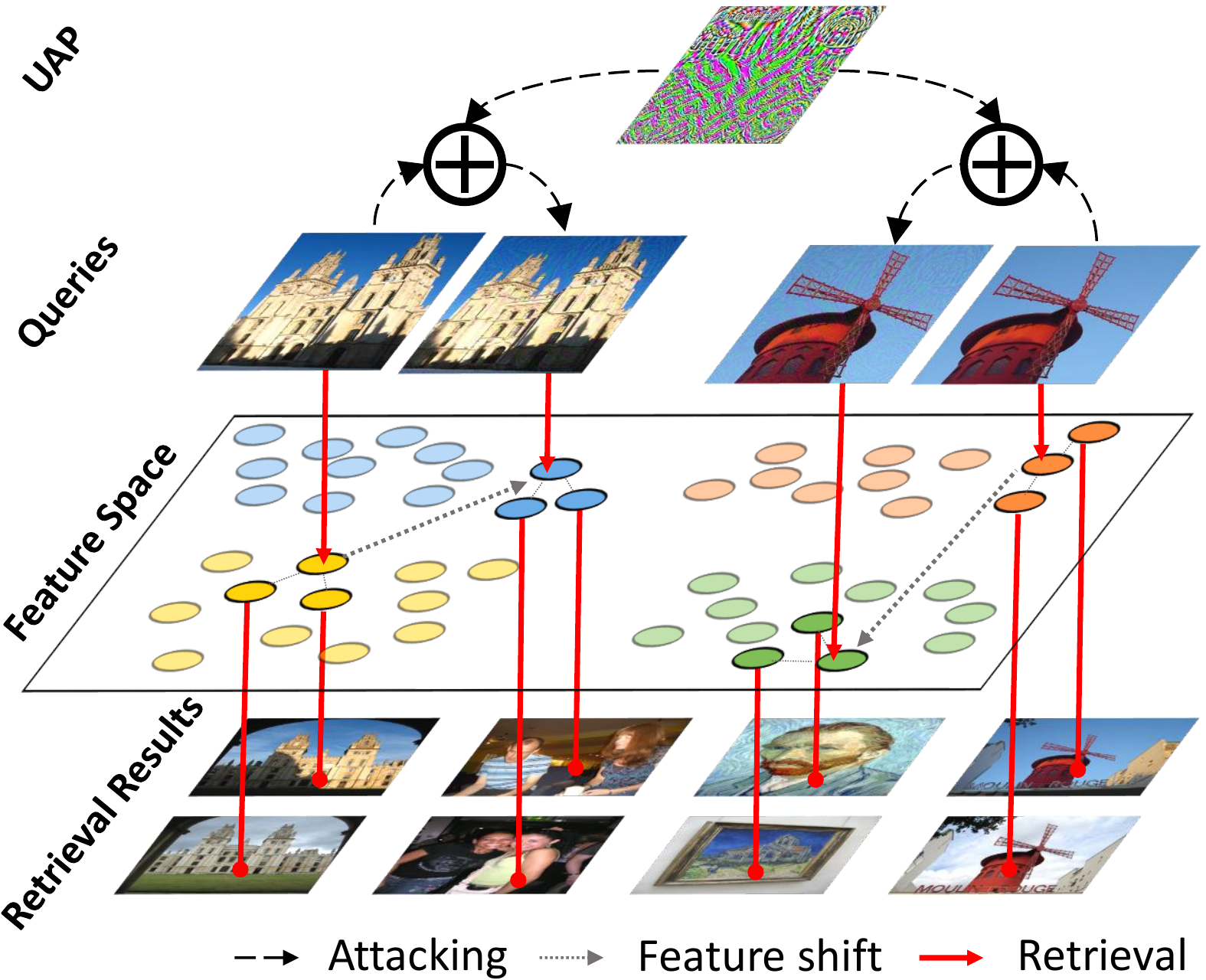}
   \caption{The UAP when added to multiple images cause most of the images to shift in the feature space without preserving original neighborhood relationships. The top is the perturbation and dots represent the features of images~\cite{uap_image_retrieval} (Best viewed in color).}
\label{fig:img_retrieval}
\end{figure}

Image retrieval is a well-established research topic in computer vision \cite{Zheng_2018}, which aims to find relevant images from a dataset given a query image. Li \etal\cite{uap_image_retrieval} was the first to propose a universal adversarial attack against image retrieval systems (see \textbf{\autoref{fig:img_retrieval}}). Concretely, image retrieval attacks are to make the retrieval system return irrelevant images to the query at the top of the ranking list. Extensive efforts have been made to improve search accuracy, but very little work has been done around the vulnerability of state-of-the-art image retrieval systems. Although retrieval systems use CNNs as the feature extractor, it is still challenging to apply existing UAPs generation techniques in image retrieval directly because of several reasons, namely (i) different dataset label formats, (ii) different goals, (iii) different sizes of model input, and (iv) different model output and optimization methods, as explained in \cite{uap_image_retrieval}.

A novel universal adversarial perturbation attack method for image retrieval is proposed in~\cite{uap_image_retrieval}, tackling the above challenges. A neural network generates a UAP, which, when used to slightly alter the query along with randomized resizing of images and UAPs, breaks the neighborhood relationships among image feature points, causing degradation in the corresponding ranking metric. Image retrieval can be viewed as a ranking problem, from which perspective the relationship between query and references plays an important role. Thus, these relationships are utilized to improve attack performance further. Such attacks showed its efficacy against the real-world visual
search engines like Google Images, revealing threats to such image-retrieval systems.

\subsection{Text Classification}

Before talking about perturbations in text classification, we first need to define what fooling a text classification system means. A perturbation in the text classification system means replacing or adding a few words with their synonymous words (see \autoref{tab:2}).~\cite{uap_text,uap_text_class} describes a token-agnostic perturbation, which when applied to each token of the text (sentence), can misclassify the text. 

\begin{table}[h]
\centering
\begin{tabular}{|p{0.75\linewidth}|c|}
\hline
Text (top: original, bottom: adversarial) & Prediction \\
\hline
i walk \textbf{and} dana runs . & 1 \\
i walk \textbf{,} dana runs . & 0 \\
\hline
\end{tabular}
  \caption{Adversarial sample generated on CoLA evaluation data whose annotations are labeled according to grammatical correctness of sentences.}
  \label{tab:2}
\end{table}

\cite{uap_text} describes a method to add a UAP to the embedding vector of the tokens of the sentence. These perturbations are generated so that the actual meaning of the sentence does not change. A norm in the n-dimensional embedding space is defined for this purpose, and then a method similar to~\cite{uap_paper} is used to generate the perturbation. There are possibilities of such attacks in NLP systems such as language translation and sentiment classification too.

\section{Future Directions}
\label{sec:future_directions}
The performance of universal adversarial perturbations has increased not only in terms of increased fooling rates and lesser or no data requirement but also in terms of convergence time required to craft such perturbations. Exploration and creation of such UAPs using different novel methods not only lay the path for better defense mechanisms but also helps us to have a better understanding of the behavior of the decision boundary of target models. \cite{uap_texture_shape} has shown that models produced by shape-biased training (trained on object shapes) are as vulnerable to UAPs as those by texture-biased (trained on image textures) ones. However, those produced by both training biases are better in performance than other model architectures. The same paper shows that untargeted UAPs are more likely to tilt the decision of the classifier towards specific class labels.

While the extensions covered in \autoref{sec:applications_of_uaps} show the applicability and effectiveness of universal adversarial perturbations in a wide variety of tasks, actual applicability of UAPs to fool modern AI systems is still far-fetched. This is mainly because the fooling rates achieved with UAPs are still very less than those achieved by per-instance perturbation generation methods~\cite{deepfool,cnw_attack}. Nevertheless, due to the minimum computation required during deployment (only the addition of UAP to the input image), UAPs are more practical for attacking any system in realtime. Hence, an increase in the fooling rates of UAPs to levels comparable to those of per-instance perturbations would further increase their applicability in the real world. This would, therefore, make way for better defense mechanisms for attacks against deep neural networks.




\section{Conclusion}

This paper presented the first comprehensive survey in the direction of universal adversarial perturbations in deep learning. Deep neural networks are found to be vulnerable to adversarial attacks regardless of their high performance and accuracy. Since \cite{uap_paper}, many papers have introduced various data-driven and data-independent algorithms to generate universal adversarial perturbations. While being quasi-imperceptible, these perturbations are transferable across multiple networks. Data-independent approaches enable attackers to fabricate white-box attacks and can pose a severe threat to security-critical applications when applied in the real world. In this paper, we surveyed various pre-eminent attacks and defense techniques for both non-targeted and targeted attacks and discussed the reason for their existence. We also discussed the application of these UAPs for various tasks such as object recognition, semantic segmentation, depth estimation, image retrieval, and text classification.

After the review, we conclude that the universal adversarial perturbations pose a significant risk to the application of deep neural networks in the physical world. The current performance of defense techniques for creating robust models is competent, but there is no elixir as such. However, owing to the active research in this field, it is hoped that various attacks and defense techniques to create robust models through deep learning will show up in the future.

\subsection*{Acknowledgments}

We thank the invaluable inputs and suggestions by Dakshit Agrawal, Aarush Gupta, and other members of Vision and Language Group, IIT Roorkee, that were integral for the successful completion of this paper. We also thank the Institute Computer Centre (ICC), IIT Roorkee, for providing GPU workstations required for performing various experiments involved in the paper.

\label{sec:conclusion}
{\small
\bibliographystyle{ieee_fullname}
\bibliography{egpaper_final}
}

\end{document}